\documentclass[sigconf,nonacm]{acmart}
\setkeys{acmart.cls}{balance=false}

\usepackage{amsmath}
\usepackage{multirow}
\usepackage{algorithm}
\usepackage{algpseudocode}
\usepackage{placeins}
\usepackage{stfloats}
\usepackage{needspace}

\setcopyright{none}
\acmDOI{}
\acmISBN{}
\date{}
\renewcommand{\shortauthors}{Pan, Xia, and Chan}
\AtBeginDocument{%
  \fancypagestyle{arxivpreprint}{%
    \fancyhf{}%
    \fancyfoot[C]{\footnotesize\thepage}%
  }%
  \pagestyle{arxivpreprint}%
}

\title[Agent-Guided Evidence Acquisition for ECPE]{Agent-Guided Evidence Acquisition for\texorpdfstring{\\}{ }Emotion-Cause Pair Extraction}

\author{Zhuangzhuang Pan}
\affiliation{%
  \institution{Universiti Malaya}
  \city{Kuala Lumpur}
  \country{Malaysia}
}
\email{23078403@siswa.um.edu.my}

\author{Yan Xia}
\affiliation{%
  \institution{Suzhou University of Technology}
  \city{Suzhou}
  \country{China}
}
\affiliation{%
  \institution{Universiti Malaya}
  \city{Kuala Lumpur}
  \country{Malaysia}
}
\email{cnxiay@szut.edu.cn}

\author{Chee Seng Chan}
\affiliation{%
  \institution{Universiti Malaya}
  \city{Kuala Lumpur}
  \country{Malaysia}
}
\affiliation{%
  \institution{VinUniversity}
  \city{Hanoi}
  \country{Vietnam}
}
\email{cs.chan@um.edu.my}

\makeatletter
\renewcommand{\@mkauthors@iii}{%
  \global\setbox\mktitle@bx=\vbox{%
    \unvbox\mktitle@bx
    \centering
    {\fontsize{12}{15}\selectfont
      Zhuangzhuang Pan\textsuperscript{1,\textdagger}\qquad
      Yan Xia\textsuperscript{2,1,\textdagger}\qquad
      Chee Seng Chan\textsuperscript{1,3,\textasteriskcentered}\par}%
    \smallskip
    {\fontsize{9}{12}\selectfont
      \textsuperscript{1}Universiti Malaya, Malaysia\quad
      \textsuperscript{2}Suzhou University of Technology, China\quad
      \textsuperscript{3}VinUniversity, Vietnam\par}%
    \smallskip
    {\footnotesize
      \textsuperscript{\textdagger}Equal contribution; shared first authorship.\quad
      \textsuperscript{\textasteriskcentered}Corresponding author: \href{mailto:cs.chan@um.edu.my}{cs.chan@um.edu.my}\par}%
    \medskip
  }%
}
\makeatother

\begin{abstract}
Emotion-Cause Pair Extraction (ECPE) requires identifying the causes associated with emotions expressed in a conversation. Accurate extraction can depend on evidence that is not initially available, making it important to decide when further information is worth acquiring. We study agent-guided evidence acquisition for ECPE, where a student agent selectively accesses expert judgments for borderline relations on both sides of its validation-selected F1 decision boundary and revises only queried predictions. Through controlled comparisons under different information-access conditions, we examine how delegation granularity, query selection, and response integration affect prediction quality. At a nominal 10\% query rate, selective fusion improves mean pair F1 by 2.64--3.88 points over the student in our main experiments, reaching 65.14 on ECF and 49.74 on MEC4 across model families. Our analysis suggests that greater access does not always translate into better cause identification. These findings motivate acquisition strategies that consider both the relevance of newly available evidence and its effect on prediction revision, providing an empirical basis for more effective and efficient evidence acquisition in ECPE. Code will be released upon acceptance.
\end{abstract}


\begin{document}
\maketitle
\thispagestyle{arxivpreprint}
\hypersetup{%
  pdfauthor={Zhuangzhuang Pan, Yan Xia, Chee Seng Chan},
  pdfkeywords={ECPE, evidence acquisition, agent delegation, query selection}%
}

\begin{figure}[!t]
\centering
\includegraphics[width=\linewidth]{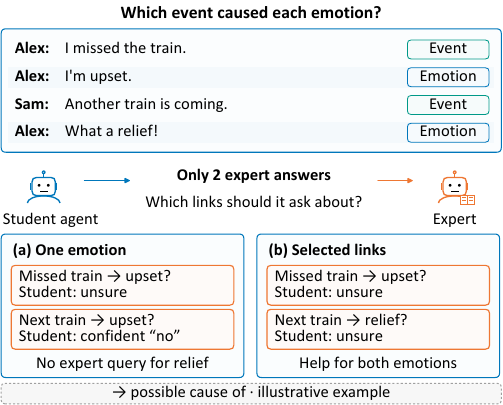}
\caption{Query allocation with two event candidates. Each orange card uses one expert answer, and status lines show the student's judgment before querying. Target delegation spends both answers on one emotion, while relation selection distributes them across emotions. The dialogue and uncertainty are illustrative.}
\Description{A dialogue links a missed train to feeling upset and an approaching train to relief. With two expert answers, querying both displayed candidates for the upset target leaves relief without expert help. Selecting uncertain relations instead distributes queries across both emotion targets.}
\label{fig:motivation}
\end{figure}

\section{Introduction}
\label{sec:introduction}

Emotion--cause pair extraction (ECPE) identifies utterance pairs linking emotions to their causes in conversations~\cite{XIA2027101055,xia2019ecpe,wang2023ecf}. These relations inform empathetic responses~\cite{he2025ecc} and connect reactions to earlier events, multiple causes, or later explanations~\cite{pan2026samedirectcausesgrounded}. This study considers a student model with limited expert access. Assistance can address a specific relation, all candidate causes of a target, or an entire dialogue.

Grouping requests by target creates an opportunity cost (Figure~\ref{fig:motivation}). An answer spent on a confidently judged relation is unavailable for an uncertain relation elsewhere in the dialogue. Delegation by relation lets the student distribute its allowance across targets, but an expert can still repeat a student error or overturn a correct judgment. Choosing when to seek help must therefore account for its likely benefit~\cite{bachar2026predictors}. We compare delegation units and query selection while holding the student, expert, response rule, and answer budget fixed.

Learning to defer formalizes decision risk given an expert's capabilities~\cite{mozannar2020defer,mao2025routing}. Language model cascades and learned routers allocate calls by predicted quality and cost~\cite{aggarwal2024automix,dekoninck2025routing,ong2025routellm}. In ECPE, a target can have several candidate causes with different levels of uncertainty. Pair F1 may also favor a threshold different from probability $0.5$~\cite{lipton2014thresholding}. Both sides of that threshold contain errors, with missed causes among rejected relations and false explanations among accepted ones. Boundary re-examination already recovers rejected ECPE candidates~\cite{feng2026judgement}.

The proposed policy uses distance from a decision threshold fixed on validation to select queries. The student agent incorporates expert judgments only at those positions. Fusion and direct replacement test the same allocation under different response rules. Entropy rankings, learned revision values, one-sided queries, and coarser delegation provide comparisons with matched expert access.

Experiments span two languages, two model families, and three independent student seeds per setting, evaluating both pair F1 and complete cause-set recovery.

The contributions are:
\begin{itemize}
\item We formulate selective expert consultation for ECPE, separating the unit of delegation, query allocation, and response integration while retaining every unqueried judgment in the student's output.
\item We show that allocation near the decision boundary competes with learned routing, improves on entropy and one-sided selection, and outperforms target or dialogue delegation with a common pairwise expert and matched access.
\item We measure net revision gains across bands around the decision boundary under fixed fusion and test consultation across response rules, expert families, calibration samples, and student training conditions.
\end{itemize}

\section{Related Work}
\label{sec:related}

\paragraph{Emotion--cause relation extraction.}
ECPE jointly identifies emotion and cause utterances~\cite{xia2019ecpe}, with conversational settings adding speaker interactions and multimodal evidence~\cite{wang2024semeval}. MECPE-2steps detects these utterances before filtering candidate pairs~\cite{wang2023ecf}. HiLo combines holistic interaction with label constraints~\cite{li2025hilo}, while M3F introduces Chinese MEC$^4$ with a memory-inspired framework~\cite{wu2025mec4}. M$^3$HG models emotion and cause contexts with heterogeneous graphs~\cite{liang2025m3hg}, and SCALE separates their semantics before alignment~\cite{ma2026scale}. BridgeProt-SFT generates dialogue-level pair sets and locally verifies causes for each proposed emotion turn, using a generative objective rather than thresholded pair classification~\cite{xia2026bridgeprot}. ECFlow combines appraisal-guided generation with graph rewards for triplet consistency~\cite{liang2026ecflow}. HCPRA organizes agents through hierarchical cognition, perception, and reasoning~\cite{wang2026hcpra}, while DEC-Debate resolves model disagreements over emotions and pairs~\cite{wu2026debate}.

\paragraph{Pair judgement and verification.}
ECPE-2D learns joint representations for pair classification~\cite{ding2020ecpe2d}, while PRG-MoE uses relation-guided classifiers for conversational pairs~\cite{jeong2023prgmoe}. GMEC supplements generative extraction with GPT-3.5 pair verification~\cite{ju2025gmec}. Feng et al.~\cite{feng2026judgement} compare generation, target selection, and pair judgement, showing that explicit pair queries recover relations omitted by generation. Their auxiliary retriever revisits the highest-ranked rejected candidate for each target. These studies motivate the pairwise foundation used here. The present study examines how to allocate limited expert access over these decisions.

\paragraph{Delegation and model cascades.}
Learning to defer accounts for expert capability~\cite{mozannar2020defer}, including consistency guarantees for routing to multiple experts~\cite{mao2025routing}. AutoMix relies on self-verification~\cite{aggarwal2024automix}, RouteLLM learns from preference data~\cite{ong2025routellm}, and GraphRouter models task--query--model interactions~\cite{feng2025graphrouter}. Unified routing and cascading estimate quality and cost~\cite{dekoninck2025routing}. Conformal Arbitrage controls risk across competing objectives~\cite{overman2025arbitrage}, and SafeRoute selectively invokes a larger safety guard~\cite{lee2025saferoute}. Allocation can also operate below the level of a whole answer. CITER routes individual generated tokens~\cite{zheng2025citer}, while selective depth reading allocates sentiment computation per aspect~\cite{xia-etal-2026-single}. The unit studied here is an independently scored emotion--cause relation within a dialogue.

\paragraph{Agent decisions and external assistance.}
Router-R1 learns when to invoke models over multiple rounds and how to aggregate their responses~\cite{zhang2025routerr1}. RouterHGC selects collaboration modes, roles, and model configurations~\cite{xiao2026routerhgc}. In human--AI moderation, performance predictors use model outputs to decide when to escalate~\cite{bachar2026predictors}. Question answering with tools must also account for uncertainty in the returned information~\cite{lymperopoulos2025tools}. We likewise separate decisions about requesting assistance from decisions about integrating the response.

\paragraph{Calibration and task adaptation.}
Calibration relates confidence to observed outcomes~\cite{guo2017calibration}, while F1 maximization determines a threshold for the task~\cite{lipton2014thresholding}. Models receive low-rank adaptation (LoRA)~\cite{hu2022lora}. A secondary knowledge distillation (KD) condition~\cite{hinton2015distilling} tests consultation after additional training with the same expert.

\begin{figure*}[!t]
\centering
\includegraphics[width=\textwidth]{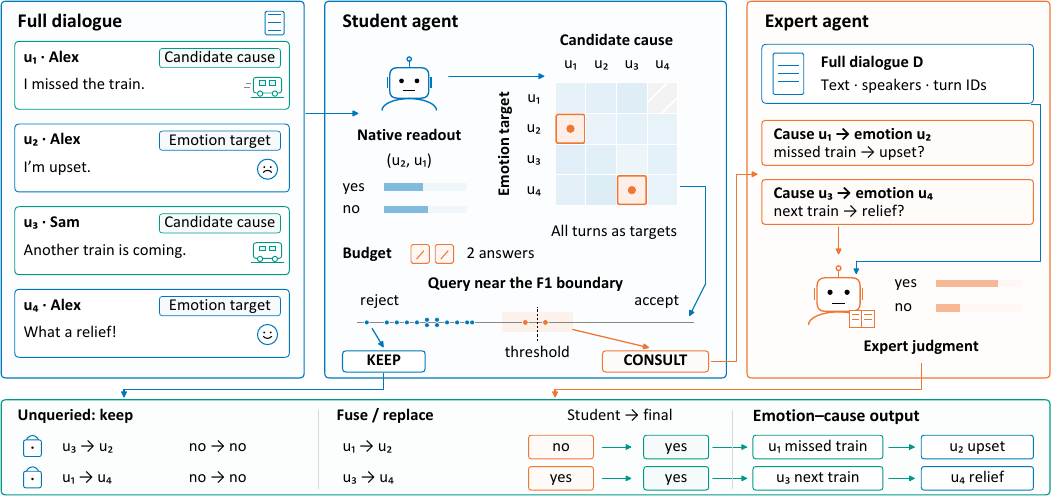}
\caption{Selective expert consultation over emotion--cause relations. The student scores every eligible target--candidate pair and queries near its F1 boundary fixed on validation. The expert receives the full dialogue and each selected relation. Fusion or direct replacement updates queried decisions while unqueried predictions remain unchanged. The dialogue, highlights, scores, and updates are illustrative, and every utterance remains a potential target.}
\Description{Indexed dialogue turns show candidate events and emotion expressions. The student scores every eligible emotion-target and candidate-cause relation, selects a bilateral boundary band, and chooses KEEP or CONSULT. The expert receives the full dialogue and selected questions. The lower panel preserves unqueried decisions, integrates queried judgments, and displays cause-to-emotion links.}
\label{fig:method}
\end{figure*}

\section{Selective Expert Consultation}
\label{sec:method}

The evidence acquired here is an expert's judgment about a relation in the given dialogue. The student agent first predicts the relations and then selects where to request assistance. Its state contains the relation graph, student scores, and query allowance. The expert receives the full dialogue and answers only the selected relation questions. Those answers can revise the queried predictions, as shown in Figure~\ref{fig:method}.

\subsection{Relations as the Unit of Delegation}
Let $D=(u_1,\ldots,u_n)$ contain utterance text, speaker identities, and turn indices. A relation $r=(i,j)$ states that $u_j$ triggers or explains a non-neutral emotion in $u_i$. The decision interface draws on pair classification~\cite{ding2020ecpe2d,jeong2023prgmoe} and LLM verification~\cite{ju2025gmec,feng2026judgement}. Every utterance is considered as a target, with candidate causes
\begin{equation}
\mathcal C(D)=\{(i,j):\max(1,i-5)\leq j\leq\min(n,i+2)\}.
\label{eq:candidates}
\end{equation}
This window includes five preceding utterances, the target itself, and two following utterances ($-5\leq j-i\leq2$). It restricts the candidate pairs. Both models still receive the full dialogue and the indexed target and candidate for every question, without gold emotion labels. Self-causes, later explanations, multiple causes, and targets with no cause are all allowed.

The student scores every relation in $\mathcal C(D)$. Delegation by target or dialogue requests all candidate answers in that group, whereas delegation by relation can distribute queries across targets. All three use the same pairwise expert interface. Evaluation retains every gold relation, counting causes outside the window as false negatives.

\subsection{Task Boundary and Query Policy}
For student $s$ and expert $e$, native yes/no logits at the first answer position give
\begin{equation}
\ell_m(r)=\ell_{m,\mathrm{yes}}(D,r)-\ell_{m,\mathrm{no}}(D,r),
\quad m\in\{s,e\}.
\label{eq:margin}
\end{equation}
Separate affine calibrators minimize validation logistic loss:
\begin{equation}
p_m(r)=\sigma(z_m(r)),\qquad z_m(r)=a_m\ell_m(r)+c_m.
\label{eq:calibration}
\end{equation}
The threshold $\tau_m$ maximizes validation pair F1 over a fixed grid. The independent decision is $\hat y_m(r)=\mathbb 1[z_m(r)>\tau_m]$.

The selector queries near this task-specific boundary:
\begin{equation}
d_s(r)=|z_s(r)-\tau_s|,\qquad
Q_b(D)=\{r\in\mathcal C(D):d_s(r)<\epsilon_b\},
\label{eq:queries}
\end{equation}
where $\epsilon_b$ is the validation quantile for nominal query fraction $b$. Each queried relation uses one expert answer. Actual query rates are measured over $N=\sum_D|\mathcal C(D)|$ eligible relations after applying the frozen band. A hard-budget variant selects the $\lfloor bN\rfloor$ smallest distances across the evaluation batch, with stable tie-breaking.

Affine calibration rescales distance from the task threshold. For $a_s\ne0$, it equals $|a_s|\,|\ell_s-(\tau_s-c_s)/a_s|$. Calibration aligns response scales without changing the ordering around the corresponding threshold.

\subsection{Why the Boundary Can Concentrate Value}
Under fusion with equal logit weights, write the student margin as $m=z_s-\tau_s$ and the expert adjustment as $\delta=(z_e-z_s)/2$. A prediction flip requires
\begin{equation}
|\delta|\geq |m|.
\label{eq:sensitivity}
\end{equation}
The student observes $|m|$ before querying, whereas $\delta$ is available only after receiving the expert answer. Small margins identify decisions where modest adjustments can affect the output.

Whether a change helps depends on its direction and the true relation. Consulting rejected predictions can recover causes, while consulting accepted predictions can remove false explanations. A bilateral band allows both. Counts of repairs and damage measure their net value.

\subsection{Local Response Integration}
A fixed revision mask separates allocation from integration:
\begin{equation}
\hat y_b(r)=
\begin{cases}
\hat y_s(r), & r\notin Q_b(D),\\
h(z_s(r),z_e(r)), & r\in Q_b(D).
\end{cases}
\label{eq:local}
\end{equation}
The primary fusion rule averages calibrated logits,
\begin{equation}
h_{\mathrm{fuse}}(z_s,z_e)=\mathbb 1[(z_s+z_e)/2>\tau_s],
\label{eq:fusion}
\end{equation}
and direct replacement adopts the expert decision,
\begin{equation}
h_{\mathrm{replace}}(z_s,z_e)=\mathbb 1[z_e>\tau_e].
\label{eq:replace}
\end{equation}
Both rules use identical query sets. A local integration ablation fits a weight $\alpha$ and threshold $\tau_q$ on validation, using $(1-\alpha)z_s+\alpha z_e>\tau_q$ inside the mask. The unqueried decisions in Eq.~\ref{eq:local} remain fixed under all three rules.

\subsection{Information Access}
The student chooses \textsc{Keep} or \textsc{Consult} before observing expert outputs, receives the requested answers, and returns the revised relation set (Algorithm~\ref{alg:consult}).

\begin{algorithm}[!t]
\caption{Selective consultation with local revision}
\label{alg:consult}
\begin{algorithmic}[1]
\Require Dialogue $D$, frozen models, calibrators, thresholds, band $\epsilon_b$, response rule $h$
\State Score $\mathcal C(D)$ with the student
\State $\hat y(r)\gets\mathbb 1[z_s(r)>\tau_s]$ for all $r$
\State Select $Q_b(D)$ from student scores (Eq.~\ref{eq:queries})
\For{$r\in Q_b(D)$}
  \State Obtain $z_e(r)$ and set $\hat y(r)\gets h(z_s(r),z_e(r))$
\EndFor
\State \Return $\{r\in\mathcal C(D):\hat y(r)=1\}$
\end{algorithmic}
\end{algorithm}

Evaluation replays fixed predictions under the same restriction: selection uses student information, and only queried expert responses enter the final predictions. Each revealed answer counts toward expert access.

\section{Experimental Protocol}
\label{sec:experiments}

\subsection{Data and Models}
Experiments use the established splits of English ECF~\cite{wang2023ecf} and Chinese MEC$^4$~\cite{wu2025mec4}, with 1,001/112/261 and 1,277/404/443 training/validation/test dialogues, respectively. Table~\ref{tab:data} connects candidate coverage to expert access, and Appendix~\ref{app:merged-A-1} gives all split statistics. Inputs are text, turn indices, and speaker identities. Existing pair annotations supply supervision. Micro pair F1 uses the complete gold count $G$:
\begin{equation}
F_1=\frac{2\mathrm{TP}}{G+\mathrm{TP}+\mathrm{FP}}.
\label{eq:f1}
\end{equation}
F1 is reported as a percentage, and differences as percentage points.
Emotion and cause endpoint F1 scores compare the distinct utterances at each end of predicted pairs with the corresponding full gold projections. Each utterance counts once per dialogue and endpoint type. These metrics measure utterance recovery from pair outputs, independently of emotion category classification.

\begin{table}[!t]
\centering
\caption{Test coverage and nominal 10\% queries. $G_{\mathrm{in}}/G$ is gold inside the window over all gold. Across seeds 42--44, $p_s^*=\sigma(\tau_s)$ gives the threshold range, $Q$ the mean answers, and $Q^+$ the mean percentage of queries on the accepted student side.}
\label{tab:data}
\begin{tabular*}{\linewidth}{@{\extracolsep{\fill}}lrrr@{}}
\toprule
\multicolumn{4}{l}{(a) Candidate coverage} \\
Dataset & Candidates $N$ & $G_{\mathrm{in}}/G$ & Coverage (\%) \\
\midrule
ECF & 16,016 & 1,843/1,873 & 98.40 \\
MEC$^4$ & 40,462 & 2,084/2,112 & 98.67 \\
\end{tabular*}
\par\smallskip
\begin{tabular*}{\linewidth}{@{\extracolsep{\fill}}llrrr@{}}
\multicolumn{5}{l}{(b) Decision thresholds and query composition} \\
Dataset & Model & $p_s^*$ range & $Q$ & $Q^+$ (\%) \\
\midrule
ECF & Qwen & 0.269--0.378 & 1,490.3 & 45.34 \\
ECF & Gemma & 0.321--0.366 & 1,454.0 & 45.37 \\
MEC$^4$ & Qwen & 0.175--0.250 & 3,906.3 & 40.58 \\
MEC$^4$ & Gemma & 0.175--0.231 & 3,905.0 & 40.00 \\
\bottomrule
\end{tabular*}
\end{table}

Student--expert pairs are Qwen3.5-4B/27B and Gemma 4 E4B-it/31B-it~\cite{qwen2026model,gemma2026model}. Each model is adapted separately on each dataset using bfloat16 (BF16) LoRA, rank 16, $\alpha=32$, dropout 0.05, effective batch size eight, and initial learning rate $5\times10^{-5}$. Qwen targets \texttt{q\_proj}, \texttt{v\_proj}, and \texttt{in\_proj\_qkv}. Gemma targets \texttt{q\_proj} and \texttt{v\_proj}. Completion cross-entropy (CE) excludes input tokens. AdamW uses no warmup or weight decay. The sequence limit is 8,192 tokens. The implementation uses NVIDIA A100-SXM4 GPUs (80 GB), Python 3.12.12, PyTorch 2.10.0 with CUDA 12.8, Transformers 5.10.1, PEFT 0.18.1, and Accelerate 1.13.0.

Primary models complete one supervised fine-tuning (SFT) pass in a linear two-pass schedule. Seeds 42, 43, and 44 start independently from the pretrained student for each setting, giving 12 students. Each family has a separate expert for each dataset, all using seed 42. Training controls continue ordinary SFT for a second pass at seed 42. A secondary condition adds one KD pass from each matching SFT parent. KD combines completion CE and binary Kullback--Leibler (KL) divergence from teacher to student with equal weights, using temperature 2 and learning rate $2\times10^{-4}$.

\subsection{Prompt}
\label{sec:prompt}
The prompt below uses the full dialogue and indexed target and candidate utterances, with original text retained for both datasets. Models use native nonthinking chat templates for the yes/no logit readout in Eq.~\ref{eq:margin}. Appendix~\ref{app:merged-B} gives full serialization details.

\par\smallskip
\begingroup
\setlength{\fboxsep}{6pt}
\setlength{\fboxrule}{0.3pt}
\noindent
\fcolorbox{black!25}{black!2}{%
\begin{minipage}{\dimexpr\linewidth-2\fboxsep-2\fboxrule\relax}
\textbf{Relation-query prompt (excerpt)}\par\smallskip
\ttfamily\raggedright
Judge the relation based on the conversational evidence; relative utterance position alone is not sufficient to accept or reject a candidate.\par\smallskip
Conversation: [full indexed dialogue]\par
Target utterance: [i. speaker: text]\par
Candidate utterance: [j. speaker: text]\par\smallskip
Question: Does the candidate utterance trigger or explain a non-neutral emotion expressed in the target utterance?\par
Answer only "yes" or "no".\par
Answer:
\end{minipage}%
}%
\endgroup
\par\smallskip

\subsection{Frozen Calibration and Access}
Affine calibrators use logistic regression with inverse regularization strength $10^6$ and at most 1,000 iterations. Thresholds maximize validation F1 over $[-8,6)$ at increments of $0.05$. Query bands use nominal 5/10/20\% allowances, with 10\% primary. Calibration, query bands, and response parameters are frozen before evaluation. Table~\ref{tab:data}(b) reports actual answers from the nominal band. A hard 10\% cap instead permits exactly 1,601 answers on ECF and 4,046 on MEC$^4$.

Random selection matches the exact query count within each dialogue and averages 20 draws. A stratified variant additionally matches the class predicted by the student and four distance categories (self, previous turn, earlier turns, and future). The primary response uses Eq.~\ref{eq:fusion}. Response ablations use the same queries. Local tuning searches $\alpha\in\{0.25,0.5,0.75\}$ and the threshold grid above. A response control permutes returned scores within each dialogue's query set, averaged over 50 draws.

\begin{table*}[!t]
\centering
\caption{Test pair F1, mean$_{\scriptscriptstyle\pm\mathrm{SD}}$ over three SFT seeds. Selective F (fusion) and Selective R (replacement) both use nominal 10\% access. Random averages 20 draws with query counts matched in each dialogue. Row maxima are bold.}
\label{tab:main}
\begin{tabular}{llrrrrrrr}
\toprule
Dataset & Model & Actual query \% & Student ($\approx$4B) & Random F & Selective F & Random R & Selective R & Expert ($\geq$27B) \\
\midrule
ECF & Qwen & 9.31 & $62.05_{\scriptscriptstyle\pm0.77}$ & $62.41_{\scriptscriptstyle\pm0.79}$ & $64.89_{\scriptscriptstyle\pm0.79}$ & $62.52_{\scriptscriptstyle\pm0.66}$ & $65.40_{\scriptscriptstyle\pm0.64}$ & $\mathbf{66.60}$ \\
ECF & Gemma & 9.08 & $61.26_{\scriptscriptstyle\pm0.66}$ & $61.79_{\scriptscriptstyle\pm0.59}$ & $65.14_{\scriptscriptstyle\pm0.29}$ & $61.86_{\scriptscriptstyle\pm0.56}$ & $65.21_{\scriptscriptstyle\pm0.05}$ & $\mathbf{66.70}$ \\
MEC$^4$ & Qwen & 9.65 & $47.10_{\scriptscriptstyle\pm0.64}$ & $47.44_{\scriptscriptstyle\pm0.63}$ & $49.74_{\scriptscriptstyle\pm0.60}$ & $47.46_{\scriptscriptstyle\pm0.61}$ & $\mathbf{49.90}_{\scriptscriptstyle\pm0.31}$ & 49.79 \\
MEC$^4$ & Gemma & 9.65 & $45.50_{\scriptscriptstyle\pm0.31}$ & $45.92_{\scriptscriptstyle\pm0.16}$ & $49.02_{\scriptscriptstyle\pm0.86}$ & $45.95_{\scriptscriptstyle\pm0.25}$ & $49.60_{\scriptscriptstyle\pm0.36}$ & $\mathbf{50.27}$ \\
\bottomrule
\relax

\end{tabular}
\end{table*}

\subsection{Competing Allocation Policies}
Native and calibrated binary entropy rank uncertainty around probability $0.5$. One-sided selectors rank distance to $\tau_s$ among either rejected or accepted predictions. A further selector admits at most one rejected candidate per target, choosing the nearest to the threshold and adapting the boundary review rule in~\cite{feng2026judgement}. When a restricted selector has too few eligible candidates, its boundary comparator uses the same reduced query count in each dialogue.

The learned router predicts revision value from 16 features available to the student. These include scores, threshold distance, predicted class, turn distance, speaker match, text lengths, target position, dialogue length, and score statistics within each target. Gradient boosting uses 120 iterations, at most seven leaves, minimum leaf size 40, learning rate 0.05, and $L_2$ regularization 1. Five validation folds split by dialogue compare reduction in logistic loss with F1-weighted count improvement as targets. Both targets are defined for each response rule. Out-of-fold F1 selects the target per allowance and access constraint. The router is then refitted on all validation dialogues. Expert outputs supply training targets only. Appendix~\ref{app:merged-E} gives the features, loss definitions, and selected configurations.

Two access constraints separate selection within dialogues from allocation across them. The first matches the boundary policy's query count in every dialogue. The second gives each selector a global cap of $\lfloor bN\rfloor$ answers, allowing it to distribute queries across dialogues. Entropy, Boundary, and Learned use identical models and response rules in both comparisons. The global 10\% comparison also refits the learned router for the continued SFT students at seed 42 using the same features and validation procedure.

Granularity comparisons use a global cap of $\lfloor bN\rfloor$ expert answers. Groups are ranked by minimum boundary distance or mean $\sigma(-d_s)$, selected on validation. A group is admitted only if all its candidate answers fit the remaining allowance. The relation comparator is also evaluated at the group's exact realized answer count.

\subsection{Uncertainty and Robustness}
Unless noted otherwise, tables report means and sample standard deviations (SD) across three student seeds. Paired 95\% bootstrap intervals use 2,000 dialogue resamples, retaining all predictions and full gold counts. Group comparisons average differences across seeds within the same resample. The intervals measure variation across dialogues for the trained models and fitted policies. SD measures variation across training seeds.

Comparisons across families exchange the expert while preserving student queries. Calibration sensitivity uses 200 resamples of validation dialogues per student, refitting maps, thresholds, and query bands. Identity checks count unqueried changes and reproduce the student at zero queries or with the student itself as expert (Appendix~\ref{app:merged-M}). Evaluation of multiple causes measures exact recovery of the full annotated cause set, including gold outside the window.

\begin{table}[!t]
\centering
\caption{Pair F1 under different backbones and source protocols. T/A/V denote text/audio/video. MECPE-2steps and HiLo follow~\cite{wu2025mec4}, and other baselines follow their original reports. Consultation uses nominal 10\% access (three-seed mean$_{\scriptscriptstyle\pm\mathrm{SD}}$). Dataset maxima are bold, and dashes denote unreported results.}
\label{tab:published}
\begin{tabular}{lcrr}
\toprule
Method & Input & ECF & MEC$^4$ \\
\midrule
MECPE-2steps~\cite{wang2023ecf} & T & 52.71 & 25.31 \\
HiLo~\cite{li2025hilo} & T & 53.09 & 34.22 \\
M3F~\cite{wu2025mec4} & T & 56.52 & 43.21 \\
SCALE~\cite{ma2026scale} & T & 57.70 & -- \\
BridgeProt-SFT~\cite{xia2026bridgeprot} & T & 53.51 & 40.53 \\
Jud-Local~\cite{feng2026judgement} & T & 62.14 & -- \\
\quad + Retriever & T & 63.60 & -- \\
\addlinespace
MECPE-2steps~\cite{wang2023ecf} & TAV & 53.20 & 28.82 \\
HiLo~\cite{li2025hilo} & TAV & 55.45 & 35.11 \\
M3F~\cite{wu2025mec4} & TAV & 55.96 & 44.79 \\
SCALE~\cite{ma2026scale} & TAV & 58.63 & -- \\
\addlinespace
Qwen selective fusion & T & $64.89_{\scriptscriptstyle\pm0.79}$ & $\mathbf{49.74}_{\scriptscriptstyle\pm0.60}$ \\
Gemma selective fusion & T & $\mathbf{65.14}_{\scriptscriptstyle\pm0.29}$ & $49.02_{\scriptscriptstyle\pm0.86}$ \\
\bottomrule
\relax

\end{tabular}
\end{table}

\begin{table*}[!t]
\centering
\caption{Query allocation at nominal 10\% access with fixed fusion and dialogue counts matched per comparison. (a) Test F1, mean$_{\scriptscriptstyle\pm\mathrm{SD}}$, and Boundary minus Learned. $H_\ell,H_p$ denote native and calibrated entropy. Best F1 is bold. (b) Bilateral minus each restricted selector. Negative/positive denote rejected/accepted student predictions. Differences include pointwise 95\% paired intervals from dialogue resampling.}
\label{tab:information}
\begin{tabular*}{\textwidth}{@{\extracolsep{\fill}}llrrrrr@{}}
\toprule
\multicolumn{7}{l}{(a) Query ranking: pair F1 and paired difference} \\
Dataset & Model & $H_\ell$ & $H_p$ & Learned & Boundary & $\Delta$ vs.\ Learned [95\% CI] \\
\midrule
ECF & Qwen & $64.09_{\scriptscriptstyle\pm0.59}$ & $63.58_{\scriptscriptstyle\pm0.60}$ & $64.47_{\scriptscriptstyle\pm0.50}$ & $\mathbf{64.89}_{\scriptscriptstyle\pm0.79}$ & $+0.42\ [ +0.04,+0.77 ]$ \\
ECF & Gemma & $64.69_{\scriptscriptstyle\pm0.67}$ & $64.18_{\scriptscriptstyle\pm0.38}$ & $64.68_{\scriptscriptstyle\pm0.50}$ & $\mathbf{65.14}_{\scriptscriptstyle\pm0.29}$ & $+0.47\ [ +0.14,+0.81 ]$ \\
MEC$^4$ & Qwen & $49.40_{\scriptscriptstyle\pm0.96}$ & $49.41_{\scriptscriptstyle\pm0.96}$ & $49.47_{\scriptscriptstyle\pm0.45}$ & $\mathbf{49.74}_{\scriptscriptstyle\pm0.60}$ & $+0.27\ [ -0.02,+0.56 ]$ \\
MEC$^4$ & Gemma & $48.64_{\scriptscriptstyle\pm0.92}$ & $48.59_{\scriptscriptstyle\pm0.82}$ & $48.56_{\scriptscriptstyle\pm0.79}$ & $\mathbf{49.02}_{\scriptscriptstyle\pm0.86}$ & $+0.46\ [ +0.21,+0.72 ]$ \\
\end{tabular*}
\par\smallskip
\begin{tabular*}{\textwidth}{@{\extracolsep{\fill}}llrrr@{}}
\multicolumn{5}{l}{(b) Query side: bilateral advantage in F1 [95\% CI]} \\
Dataset & Model & Negative only & Positive only & One rejected/target \\
\midrule
ECF & Qwen & $+1.21\ [ +0.76,+1.61 ]$ & $+1.09\ [ +0.54,+1.62 ]$ & $+1.24\ [ +0.79,+1.66 ]$ \\
ECF & Gemma & $+1.10\ [ +0.50,+1.72 ]$ & $+2.03\ [ +1.36,+2.66 ]$ & $+1.69\ [ +1.07,+2.30 ]$ \\
MEC$^4$ & Qwen & $+1.24\ [ +0.67,+1.77 ]$ & $+0.84\ [ +0.33,+1.38 ]$ & $+1.33\ [ +0.76,+1.89 ]$ \\
MEC$^4$ & Gemma & $+1.51\ [ +0.97,+2.03 ]$ & $+1.08\ [ +0.54,+1.64 ]$ & $+1.65\ [ +1.11,+2.22 ]$ \\
\bottomrule
\end{tabular*}
\end{table*}

\section{Results and Analysis}
\label{sec:results}

\subsection{Selective Allocation Improves Extraction}
Selective querying makes better use of a fixed expert budget (Table~\ref{tab:main}). Random consultation changes student F1 only modestly, while selective fusion gains 2.64--3.88 points over the student with query counts matched in each dialogue. Every student's paired interval favors selective over random fusion (Appendix~\ref{app:merged-C-2}). With the expert, access, and response rule fixed, the difference comes from query selection. The advantage also holds under direct replacement.

Matching random queries by predicted class and relation distance, as well as dialogue, retains a mean selective advantage of 1.41--2.02 F1 points across settings. Intervals are positive for all twelve students (Appendix~\ref{app:merged-F-4}). Boundary distance therefore identifies useful queries within the same class and position groups, beyond any benefit from requesting more predicted positives or nearby utterances.

Table~\ref{tab:published} places these results alongside encoder and decoder methods with different backbones, candidate scopes, and training protocols. Selective fusion attains the highest listed F1 on both datasets. Jud-Local and its Retriever extension~\cite{feng2026judgement} provide a close design reference. The present selector reuses native logits with no additional neural forward pass, replacing retriever-specific fine-tuning with a separately adapted large expert and its storage and inference costs. The matched comparisons in Table~\ref{tab:main} isolate allocation effects.

Consultation improves identification of emotion and cause utterances as well as their pairing. Both fusion and replacement raise emotion and cause endpoint F1 in every setting (Appendix~\ref{app:merged-C-3}).

\subsection{Query Center and Sides Matter}
F1 thresholds below probability $0.5$ shift the relevant uncertainty region (Table~\ref{tab:data}(b)). With dialogue query counts fixed, boundary ranking outperforms entropy at nominal 10\% access (Table~\ref{tab:information}(a)). Selection benefits from locating decisions near the threshold actually used for extraction.

Boundary is competitive with Learned, whose extra features offer no consistent allocation advantage. Paired intervals favor Boundary in three settings and include zero for Qwen/MEC$^4$. Its mean advantage at 10\% also holds under global allocation and direct replacement (Appendix~\ref{app:merged-E}). The gain varies with the allowance, with Qwen/ECF nearly tied at 5\%.

Both sides of the boundary receive substantial access (Table~\ref{tab:data}(b)). Bilateral selection outperforms all restricted selectors at matched actual counts (Table~\ref{tab:information}(b)), allowing expert help to address missed causes and false explanations together.

\subsection{Grouping Queries Has an Opportunity Cost}
Delegating a whole group ties a useful query to other relations the student might not choose individually. Under the common pairwise expert interface, relation selection yields higher F1 than target or dialogue selection (Table~\ref{tab:granularity}). Its advantage persists when its query count is reduced to the exact number used by each group policy (Appendix~\ref{app:merged-D-1}). Grouping therefore affects the value of the queried set even after accounting for unused budget.

\begin{table}[!t]
\centering
\caption{Delegation under a hard 10\% answer cap with fixed fusion, mean$_{\scriptscriptstyle\pm\mathrm{SD}}$ across three seeds. (a) Test pair F1. (b) Exact cause-set recovery (\%) for targets with at least two gold causes, with 417 on ECF and 218 on MEC$^4$. Row maxima are bold.}
\label{tab:granularity}
\begin{tabular}{@{}llrrr@{}}
\toprule
\multicolumn{5}{l}{(a) Pair F1} \\
Dataset & Model & Relation & Target & Dialogue \\
\midrule
ECF & Qwen & $\mathbf{64.95}_{\scriptscriptstyle\pm0.81}$ & $63.08_{\scriptscriptstyle\pm0.63}$ & $62.70_{\scriptscriptstyle\pm1.02}$ \\
ECF & Gemma & $\mathbf{65.22}_{\scriptscriptstyle\pm0.47}$ & $62.71_{\scriptscriptstyle\pm0.50}$ & $61.86_{\scriptscriptstyle\pm0.93}$ \\
MEC$^4$ & Qwen & $\mathbf{49.77}_{\scriptscriptstyle\pm0.59}$ & $47.97_{\scriptscriptstyle\pm0.60}$ & $47.51_{\scriptscriptstyle\pm0.60}$ \\
MEC$^4$ & Gemma & $\mathbf{49.06}_{\scriptscriptstyle\pm0.87}$ & $46.55_{\scriptscriptstyle\pm0.44}$ & $45.74_{\scriptscriptstyle\pm0.28}$ \\
\addlinespace
\multicolumn{5}{l}{(b) Complete multi-cause sets (\%)} \\
Dataset & Model & Relation & Target & Dialogue \\
\midrule
ECF & Qwen & $\mathbf{38.85}_{\scriptscriptstyle\pm6.17}$ & $35.73_{\scriptscriptstyle\pm4.99}$ & $35.09_{\scriptscriptstyle\pm5.48}$ \\
ECF & Gemma & $\mathbf{34.69}_{\scriptscriptstyle\pm2.45}$ & $31.65_{\scriptscriptstyle\pm2.59}$ & $31.25_{\scriptscriptstyle\pm2.56}$ \\
MEC$^4$ & Qwen & $\mathbf{18.81}_{\scriptscriptstyle\pm2.75}$ & $17.28_{\scriptscriptstyle\pm1.74}$ & $16.67_{\scriptscriptstyle\pm2.26}$ \\
MEC$^4$ & Gemma & $\mathbf{17.28}_{\scriptscriptstyle\pm3.74}$ & $14.53_{\scriptscriptstyle\pm1.61}$ & $13.91_{\scriptscriptstyle\pm1.91}$ \\
\bottomrule
\end{tabular}
\end{table}

Relation selection also improves exact recovery of multi-cause sets (Table~\ref{tab:granularity}(b) and Appendix~\ref{app:merged-D-2}), extending the benefit of individual query allocation to complete sets despite independent expert judgments.

\subsection{Access, Integration, and Execution Cost}
\begin{figure*}[!t]
\centering
\includegraphics[width=\textwidth]{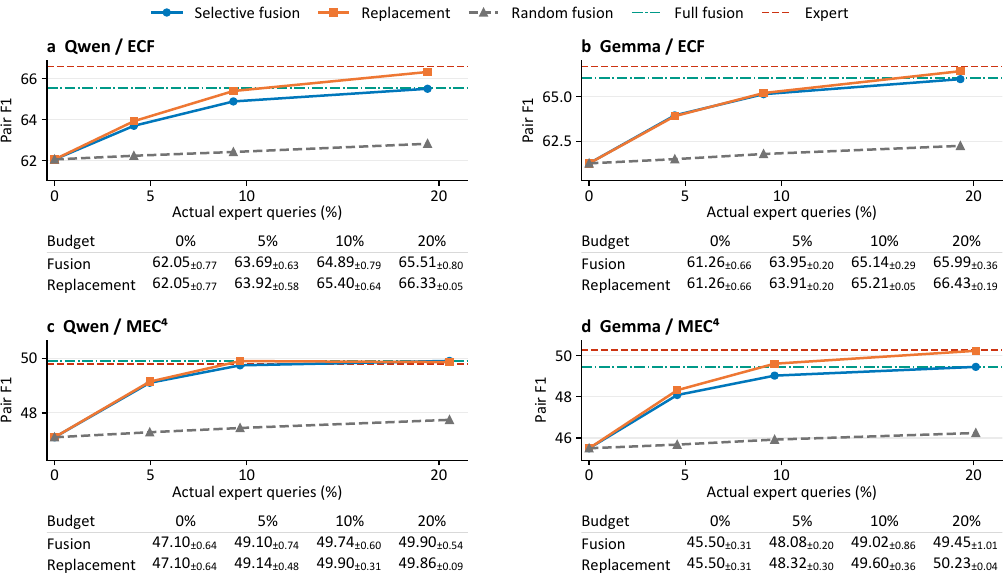}
\caption{Expert access and test pair F1 across three SFT seeds. Points show means. Inset rows report mean$_{\scriptscriptstyle\pm\mathrm{SD}}$ at nominal 0/5/10/20\% access. The horizontal references distinguish full fusion at the student threshold from the independent expert. Curves use actual query fractions. Random fusion matches query counts within every dialogue.}
\Description{Four panels plot mean F1 for selective fusion, direct replacement, and random fusion. Inset tables report means and sample standard deviations across three student seeds, including the zero-query endpoint. Separate horizontal lines mark full fusion and the expert. A single legend is above the panels.}
\label{fig:frontier}
\end{figure*}

Fusion and replacement have different returns from wider access (Figure~\ref{fig:frontier} and Appendix~\ref{app:merged-L}). Fixed fusion captures most of its full-access gain early, whereas direct replacement often continues to improve as the band widens. Averaging reduces an expert's influence when the student is confident. Replacement allows the expert to override those judgments, so answers farther from the boundary can still change the output.

At 20\% nominal access, replacement is within 0.27 F1 of the independent expert in every setting, with paired difference intervals including zero. Selective fusion and full fusion produce identical prediction sets for all six MEC$^4$ students and differ on only 0--7 relations per ECF student (Appendix~\ref{app:merged-I-2}). Their plateau reflects unchanged decisions outside the query band. Replacement can still use those expert answers to make revisions that averaging suppresses.

Fitting a weight and threshold for queried relations adds 0.38--0.83 F1 points over fixed fusion at 20\%, leaving all other predictions unchanged (Appendix~\ref{app:merged-G}). Equal logit weights can therefore underuse the selected expert answers.

\paragraph{From answer access to execution cost.}
Table~\ref{tab:timing} evaluates the policies online on 32 ECF validation dialogues with 1,847 candidate relations. Each family runs alone on one A100 after warming up on the full panel, with three repeats in rotating order. The measured interval includes scoring, selection, integration, and required model transfers. Qwen retains both models and reuses dialogue prefixes. Gemma alternates model residency and uses native independent batches.

\begin{table}[!t]
\centering
\caption{Online time in seconds, mean$_{\scriptscriptstyle\pm\mathrm{SD}}$ over three warmed repeats. Selective fusion uses nominal 10\% access. All methods process the same 1,847 relations.}
\label{tab:timing}
\begin{tabular}{@{}lrrrr@{}}
\toprule
Model & Student & Expert & Full fusion & Selective \\
\midrule
Qwen & $15.57_{\scriptscriptstyle\pm0.10}$ & $61.58_{\scriptscriptstyle\pm0.16}$ & $76.88_{\scriptscriptstyle\pm0.07}$ & $27.60_{\scriptscriptstyle\pm0.02}$ \\
Gemma & $70.49_{\scriptscriptstyle\pm0.14}$ & $338.33_{\scriptscriptstyle\pm0.38}$ & $461.67_{\scriptscriptstyle\pm1.66}$ & $154.27_{\scriptscriptstyle\pm0.11}$ \\
\bottomrule
\end{tabular}
\end{table}

Selective consultation provides a 2.79$\times$ speedup over full fusion for Qwen and a 2.99$\times$ speedup for Gemma. The queries still reach 30 and 29 of the 32 dialogues, respectively, retaining much of the dialogue processing overhead. Query savings therefore translate only partly into runtime savings. Matched random consultation takes similar time but yields lower F1. Appendix~\ref{app:merged-H} details online execution, memory use, and agreement with cached-score replay.

\subsection{Revision Value by Location and Rule}
Under fixed fusion, net correct revisions per query are highest nearest the decision boundary (Table~\ref{tab:repairs}). Scores farther from the threshold rarely cross it after averaging, so most additional answers leave predictions unchanged. This pattern agrees with the sensitivity condition in Eq.~\ref{eq:sensitivity}.

\begin{table}[!t]
\centering
\caption{Net correct revisions per 100 queries under fixed fusion, pooled across three seeds. Bins use validation distance quantiles outward from the student threshold. Row maxima are bold.}
\label{tab:repairs}
\begin{tabular}{@{}llrrrr@{}}
\toprule
Dataset & Model & 0--5\% & 5--10\% & 10--20\% & 20--100\% \\
\midrule
ECF & Qwen & $\mathbf{10.43}$ & $6.92$ & $1.32$ & $0.01$ \\
ECF & Gemma & $\mathbf{16.25}$ & $6.40$ & $1.87$ & $0.02$ \\
MEC$^4$ & Qwen & $\mathbf{5.02}$ & $2.96$ & $0.11$ & $0.00$ \\
MEC$^4$ & Gemma & $\mathbf{8.77}$ & $1.89$ & $0.47$ & $0.00$ \\
\bottomrule
\end{tabular}
\end{table}

\begin{table}[!t]
\centering
\caption{Consultation after further training with the original expert. Fixed fusion uses nominal 10\% access in (a,b) and a global 10\% cap in (c). Values are test F1, with mean$_{\scriptscriptstyle\pm\mathrm{SD}}$ in (b). $H_\ell,H_p$ denote native and calibrated entropy. Row maxima are bold.}
\label{tab:training}
\begin{tabular*}{\linewidth}{@{\extracolsep{\fill}}llrrr}
\toprule
\multicolumn{5}{l}{(a) Two ordinary SFT passes (seed 42)} \\
Dataset & Model & Student & Random & Selective \\
\midrule
ECF & Qwen & 62.66 & 63.06 & $\mathbf{65.89}$ \\
ECF & Gemma & 62.59 & 63.00 & $\mathbf{65.77}$ \\
MEC$^4$ & Qwen & 47.73 & 47.95 & $\mathbf{49.70}$ \\
MEC$^4$ & Gemma & 46.88 & 47.30 & $\mathbf{50.16}$ \\
\addlinespace
\multicolumn{5}{l}{(b) One SFT pass plus one KD pass (three seeds)} \\
Dataset & Model & Student & Random & Selective \\
\midrule
ECF & Qwen & $64.95_{\scriptscriptstyle\pm0.09}$ & $65.11_{\scriptscriptstyle\pm0.08}$ & $\mathbf{66.49}_{\scriptscriptstyle\pm0.25}$ \\
ECF & Gemma & $63.80_{\scriptscriptstyle\pm0.47}$ & $64.14_{\scriptscriptstyle\pm0.56}$ & $\mathbf{66.02}_{\scriptscriptstyle\pm1.15}$ \\
MEC$^4$ & Qwen & $49.87_{\scriptscriptstyle\pm0.05}$ & $49.92_{\scriptscriptstyle\pm0.05}$ & $\mathbf{50.36}_{\scriptscriptstyle\pm0.17}$ \\
MEC$^4$ & Gemma & $49.10_{\scriptscriptstyle\pm0.21}$ & $49.23_{\scriptscriptstyle\pm0.23}$ & $\mathbf{50.34}_{\scriptscriptstyle\pm0.31}$ \\
\end{tabular*}
\par\smallskip
\begin{tabular*}{\linewidth}{@{\extracolsep{\fill}}llrrrr}
\multicolumn{6}{l}{(c) Two SFT passes (seed 42), hard global 10\% cap} \\
Dataset & Model & $H_\ell$ & $H_p$ & Learned & Boundary \\
\midrule
ECF & Qwen & 64.40 & 64.15 & 65.16 & $\mathbf{65.98}$ \\
ECF & Gemma & 65.00 & 65.02 & 64.81 & $\mathbf{65.74}$ \\
MEC$^4$ & Qwen & 49.49 & 49.39 & 49.15 & $\mathbf{49.69}$ \\
MEC$^4$ & Gemma & 50.08 & 50.06 & 49.30 & $\mathbf{50.15}$ \\
\bottomrule
\end{tabular*}
\end{table}

Replacement can overturn confident judgments farther from the boundary, including correct ones. Some outer bands consequently have negative net value (Appendix~\ref{app:merged-I-1}). The value of a wider allowance depends on the response rule as well as query location, accounting for the diminishing or negative returns under replacement.

Repairs also have different effects on pair F1. Recovering a cause changes TP, while suppressing an incorrect explanation changes FP. Relative to baseline TP and FP, and retaining the full gold count $G$, F1 improves when
\begin{equation}
\Delta\mathrm{TP}(G+\mathrm{FP})>\mathrm{TP}\,\Delta\mathrm{FP}.
\label{eq:gain}
\end{equation}
The same number of net correct revisions can therefore produce different F1 gains, depending on how the revisions change TP and FP. The observed gains combine recovered causes with removed false positives and extend to external relations as well as self-causes (Appendix~\ref{app:merged-C-4}).

The returned answer must address the relation that was queried. Permuting expert scores within each dialogue's query set preserves the positions and score distribution but lowers F1 by 3.63--5.55 points (Appendix~\ref{app:merged-I-3}).

\subsection{Consultation Across Training and Experts}
Consultation remains useful after further ordinary SFT. Table~\ref{tab:training}(a) covers all four dataset--model settings at seed 42, retaining the original expert from the first pass. Selective fusion exceeds both the continued student and matched random consultation, with positive paired intervals in every case. Appendix~\ref{app:merged-J-1} reports the full access curves.

Boundary gains 0.54--0.93 F1 points over the refitted learned router under the global 10\% cap (Table~\ref{tab:training}(c)), with positive paired intervals in all four settings. Entropy is close on MEC$^4$, while ECF retains a clearer advantage for the task threshold. Appendix~\ref{app:merged-J-2} gives the full comparison.

In the secondary KD condition, consultation retains positive mean gains (Table~\ref{tab:training}(b)). The advantage over random access is smaller on Qwen/MEC$^4$, where the KD student already matches the expert's standalone F1. Nine of the twelve intervals against random consultation are positive, while the three Qwen/MEC$^4$ intervals overlap zero.

With student queries fixed, consultation remains useful after changing the expert family, but the effect of the swap depends on direction and dataset. Similar standalone F1 therefore need not make experts equally useful to a given student. Calibration resampling preserves the gain in all 200 validation resamples per student, despite changes to thresholds and queried relations. Appendix~\ref{app:merged-K} reports both checks.

\section{Discussion}
\label{sec:discussion}

The results motivate aligning acquisition with the deployed decision threshold and evaluating repairs together with damage. Student uncertainty guides selection, while the response rule determines which expert answers change the output. Overall expert F1 alone captures neither interaction. These considerations connect query allocation to its effect on the final task decisions.

The granularity comparison holds expert answers fixed through a common pairwise interface. Mandatory groups then spend part of the allowance on relations that individual selection would skip. A joint cause-set interface could change both answer quality and group cost. Runtime also reflects dialogue coverage, batching, prefix reuse, and model residency. Delegation should therefore be evaluated against the available response interface and its execution costs.

\paragraph{Limitations.}
The current study uses textual transcripts, leaving acoustic and visual evidence outside the consultation process. Incorporating audio and video would require deciding which modality to inspect as well as which relation to query. Experiments restrict candidate offsets to $[-5,2]$ while retaining the full dialogue as context. Relations outside this window cannot be recovered and remain false negatives. Performance and query allocation over the full $n^2$ candidate space remain untested. Future work will examine multimodal consultation and adaptive candidate expansion, with acquisition costs measured for the additional inputs.

\section{Conclusion}
\label{sec:conclusion}

Selective consultation improves ECPE by directing limited expert assistance to borderline relations. Boundary selection is competitive with learned routing across response rules and training conditions. With a common pairwise expert, relation selection outperforms target and dialogue grouping, while bilateral queries address both error directions. Revision analysis links these gains to query location and response integration.

\bibliographystyle{ACM-Reference-Format}
\bibliography{references}

\FloatBarrier
\appendix
\setcounter{table}{0}
\setcounter{figure}{0}
\setcounter{equation}{0}
\renewcommand{\thetable}{S\arabic{table}}
\renewcommand{\thefigure}{S\arabic{figure}}
\renewcommand{\theequation}{S\arabic{equation}}
\renewcommand{\theHtable}{appendix.\arabic{table}}
\renewcommand{\theHfigure}{appendix.\arabic{figure}}
\renewcommand{\theHequation}{appendix.\arabic{equation}}
\setcounter{topnumber}{4}
\setcounter{bottomnumber}{3}
\setcounter{totalnumber}{6}
\setcounter{dbltopnumber}{4}
\renewcommand{\topfraction}{0.95}
\renewcommand{\bottomfraction}{0.90}
\renewcommand{\dbltopfraction}{0.95}
\renewcommand{\textfraction}{0.05}
\renewcommand{\floatpagefraction}{0.80}
\renewcommand{\dblfloatpagefraction}{0.80}
\captionsetup[table]{skip=5pt}
\setlength{\floatsep}{10pt plus 2pt minus 2pt}
\setlength{\textfloatsep}{10pt plus 2pt minus 2pt}
\setlength{\intextsep}{10pt plus 2pt minus 2pt}
\setlength{\dblfloatsep}{10pt plus 2pt minus 2pt}
\setlength{\dbltextfloatsep}{10pt plus 2pt minus 2pt}
\makeatletter
\setlength{\@fptop}{0pt}
\setlength{\@fpsep}{10pt}
\setlength{\@fpbot}{0pt plus 1fil}
\setlength{\@dblfptop}{0pt}
\setlength{\@dblfpsep}{10pt}
\setlength{\@dblfpbot}{0pt plus 1fil}
\makeatother
\Needspace{10\baselineskip}
\section*{Appendix}
This appendix reports how selective consultation varies with student seed, query budget, delegation unit, and response rule. Further comparisons examine competing selectors, expert identity, calibration, additional student training, and online execution. The appendix also records model inputs, answer readout, training objectives, and frozen policy settings.

\section{Experimental Setting}
\label{app:merged-A}
\label{sec:s-setting}
All comparisons use the same candidate definitions and evaluation counts.

\subsection{Data, candidates, and gold coverage}
\label{app:merged-A-1}
Every policy receives the complete dialogue and the same P5/F2 candidate set. A target can pair with itself, any of its five preceding utterances, or either of its two following utterances. Table~\ref{tab:s-data} reports the resulting coverage. ECF is English and MEC$^4$ is Chinese. Inputs contain text, turn indices, and speaker identities. Existing pair annotations provide all supervision.

The fixed window excludes 30 ECF test gold relations and 28 MEC$^4$ test gold relations. These remain in the denominator as false negatives and cannot be recovered by consultation.

\begin{table}[!htbp]
\centering
\caption{Full dataset statistics for the P5/F2 candidate window. D, U, N, and G denote dialogues, utterances, candidates, and gold pairs. O counts gold pairs outside the window, retained as false negatives.}
\label{tab:s-data}
\begin{tabular}{@{}llrrrrr@{}}
\toprule
Dataset & Split & D & U & N & G & O \\
\midrule
ECF & Train & 1,001 & 9,966 & 62,604 & 7,055 & 122 \\
ECF & Validation & 112 & 1,087 & 6,760 & 866 & 17 \\
ECF & Test & 261 & 2,566 & 16,016 & 1,873 & 30 \\
MEC$^4$ & Train & 1,277 & 17,513 & 117,158 & 5,777 & 62 \\
MEC$^4$ & Validation & 404 & 5,541 & 37,085 & 1,847 & 23 \\
MEC$^4$ & Test & 443 & 6,051 & 40,462 & 2,112 & 28 \\
\bottomrule
\end{tabular}
\end{table}

\subsection{Student and expert models}
\label{app:merged-A-2}
Students are Qwen3.5-4B and Gemma 4 E4B-it. Their corresponding experts are Qwen3.5-27B and Gemma 4 31B-it. Each model is adapted separately on each dataset with bfloat16 (BF16) low-rank adaptation (LoRA). The primary checkpoint follows one supervised fine-tuning (SFT) pass. Student seeds 42, 43, and 44 start independently from the pretrained model. Each family--dataset setting uses one fixed expert trained with seed 42, shared by the three students. Section~\ref{sec:s-kd} examines ordinary SFT continuation and an additional knowledge distillation (KD) pass separately.

SFT uses rank 16, $\alpha=32$, dropout 0.05, and initial learning rate $5\times10^{-5}$. The effective batch is eight candidate pairs, the maximum sequence length is 8,192 tokens, and AdamW has no weight decay or warmup. The checkpoint preserves the first pass of a linear two-pass schedule, before its learning rate has decayed to zero. Native completion cross-entropy (CE) excludes input tokens. Qwen adapters target \texttt{q\_proj}, \texttt{v\_proj}, and \texttt{in\_proj\_qkv}. Gemma adapters target \texttt{q\_proj} and \texttt{v\_proj}.

Similar expert F1 can conceal different precision--recall profiles, particularly on MEC$^4$ (Table~\ref{tab:s-experts}). These differences may affect which expert is useful for a student's queries. Section~\ref{sec:s-robust} tests this possibility by exchanging experts while keeping the queries fixed.

\begin{table}[!htbp]
\centering
\caption{Fixed expert models. All use seed 42 and one SFT data pass. Each checkpoint is shared across the three student seeds in its setting.}
\label{tab:s-experts}
\begin{tabular}{@{}lllrrrr@{}}
\toprule
Dataset & Model & Seed & Val F1 & Test F1 & Test P & Test R \\
\midrule
ECF & Qwen & 42 & 67.42 & 66.60 & 62.06 & 71.86 \\
ECF & Gemma & 42 & 67.45 & 66.70 & 64.80 & 68.71 \\
MEC$^4$ & Qwen & 42 & 48.74 & 49.79 & 42.37 & 60.37 \\
MEC$^4$ & Gemma & 42 & 48.80 & 50.27 & 47.98 & 52.79 \\
\bottomrule
\relax

\end{tabular}
\end{table}

\subsection{Response rules and statistical reporting}
\label{app:merged-A-3}
\label{sec:s-statistics}
F denotes fusion with equal logit weights at the unchanged student threshold. R denotes direct replacement at the expert threshold. Both act only on queried relations. Random F and Random R average 20 random draws with exactly the selective policy's query count in each dialogue. A selector sees the student's scores before querying and receives expert scores only at the selected positions.

Group summaries are arithmetic means and sample standard deviations (SD) over three students. Random policy results are first averaged within each student. Paired intervals use 2,000 bootstrap resamples of dialogues, retaining every candidate decision and the complete gold count. For a comparison with random consultation, each resample averages F1 over the same 20 random draws before subtraction. These intervals describe dialogue variation conditional on the trained models and frozen calibration. The standard deviations describe variation across student training seeds.

\section{Model Input and Answer Readout}
\label{app:merged-B}
\label{sec:s-readout}
Every relation prompt contains the complete dialogue, the target utterance, and the candidate utterance. Turn indices start at one. All targets are scored, including those without a gold emotion--cause pair. Candidate indices range from $\max(1,i-5)$ through $\min(n,i+2)$, including the target itself. Each candidate supplies one training example with the annotated binary relation as its answer. All negative candidates are retained. Emotion labels are absent from the input.

The English instructions below are identical for ECF and MEC$^4$. The three bracketed blocks contain the original dialogue, target, and candidate text with their speaker identifiers, without translation or a generated rationale.

\subsection{Relation prompt}
\label{app:merged-B-1}
\label{sec:s-prompt}
Each utterance is serialized as \texttt{index. speaker: text}. The target and candidate blocks repeat the designated utterance. The complete user message is:
\begin{quote}
\ttfamily\raggedright
You are an expert in emotion--cause pair extraction in conversations.\par
Given the following conversation, determine whether the candidate utterance triggers or explains a non-neutral emotion expressed in the target utterance.\par
A valid cause may be an event, statement, behavior, or contextual clue that triggers or explains the target emotion.\par
Judge the relation based on the conversational evidence; relative utterance position alone is not sufficient to accept or reject a candidate.\par
Conversation:\par
[all utterances in chronological order]\par
Target utterance:\par
[target index. speaker: text]\par
Candidate utterance:\par
[candidate index. speaker: text]\par
Question:\par
Does the candidate utterance trigger or explain a non-neutral emotion expressed in the target utterance?\par
Answer only "yes" or "no".\par
Answer:
\end{quote}

\subsection{Native answer serialization}
\label{app:merged-B-2}

Each native chat input contains a single user message. Serialization sets \texttt{add\_generation\_prompt} to \texttt{True} and \texttt{enable\_thinking} to \texttt{False}. No separate system message is inserted. Let $L$ denote the unpadded serialized prompt length. The first answer token occupies position $L$, with zero-based indexing, and is predicted by the logits at position $L-1$. The relation score is the native yes logit minus the native no logit. No answer tokens are generated during scoring.

Table~\ref{tab:s-token} records the four tokenizer contracts. Answer IDs agree within each model family, while the native assistant prefixes differ. The following serializations locate the first answer position for each endpoint and tie its inference score to the supervised completion.

\begin{table}[!htbp]
\centering
\caption{Answer token identities from the exact local tokenizer revisions. Full hashes and serialized examples are supplied in the evidence files.}
\label{tab:s-token}
\begin{tabular}{@{}lrrrl@{}}
\toprule
Endpoint & yes & no & pad & Revision \\
\midrule
Qwen 4B & 9405 & 2083 & 248044 & \texttt{851bf6e806ef} \\
Qwen 27B & 9405 & 2083 & 248044 & \texttt{fc05daec18b0} \\
Gemma E4B & 4443 & 1904 & 0 & \texttt{ee0ef6023621} \\
Gemma 31B & 4443 & 1904 & 0 & \texttt{842da3794eaa} \\
\bottomrule
\end{tabular}
\end{table}

\paragraph{Qwen student and expert serialization.} The assistant prefix is
\begin{verbatim}
<|im_end|>
<|im_start|>assistant
<think>
</think>
\end{verbatim}

The supervised completion IDs are \texttt{[9405, 248046, 198]} for yes and \texttt{[2083, 248046, 198]} for no.

\paragraph{Gemma student serialization.} The assistant prefix is
\begin{verbatim}
<turn|>
<|turn>model
\end{verbatim}

The supervised completion IDs are \texttt{[4443, 106, 107]} for yes and \texttt{[1904, 106, 107]} for no.

\begin{samepage}
\paragraph{Gemma expert serialization.} The assistant prefix is
\begin{verbatim}
<turn|>
<|turn>model
<|channel>thought
<channel|>
\end{verbatim}

The supervised completion IDs are \texttt{[4443, 106, 107]} for yes and \texttt{[1904, 106, 107]} for no.
\end{samepage}

For Qwen, independent suffixes with right padding use their actual final token positions. Execution with a shared prefix gathers the corresponding last suffix hidden state and applies the original language model head over the full vocabulary. For Gemma, scoring with left padding reads \texttt{logits[:, -1, [yes, no]]} from the native forward pass with \texttt{logits\_to\_keep=1}. Its native output soft-capping remains active. The token records contain complete serialized example inputs and their integer token sequences for all four models.

\subsection{Training and scoring consistency}
\label{app:merged-B-3}
Training supervises the native completion, including its end marker, and masks every prompt and padding token with label $-100$. Qwen verifies that the serialized assistant answer extends the generation prefix exactly. Gemma obtains native answer suffixes from the yes/no assistant serializations and attaches them to the same nonthinking generation prefix used in inference. The native channel delimiters are present in both training and scoring. The collator verifies that the first supervised token is the designated yes or no ID. Training sequences and scoring prompts longer than 8,192 tokens raise an error instead of being truncated.

\subsection{Frozen calibration and query bands}
\label{app:merged-B-4}
\label{sec:s-calibration}
Separate affine sigmoid calibrators align student and expert validation logits. Thresholds maximize validation pair F1 over $[-8,6)$ at increments of 0.05. Query distance is measured from the student's threshold in calibrated logit units. Table~\ref{tab:s-calibration} records the resulting thresholds and bands. The accompanying JSON retains full precision and both affine coefficients. The same settings are applied to test.

The thresholds differ across students and datasets, so centering queries at probability $0.5$ would select different relations. Band widths also vary by student. A relation is queried only when its distance is strictly below the band width, making the actual fraction sensitive to tied scores and shifts in the score distribution. Comparisons with a hard budget instead select exactly $\lfloor bN\rfloor$ relations by rank.

\begin{table*}[!tbp]
\centering
\caption{Frozen logit thresholds and query bands. $\tau_f$ is selected separately on validation for full fusion. The fixed-fusion access curves instead retain $\tau_s$ at every query rate, including full access. Bands are rounded here, with full precision supplied in JSON.}
\label{tab:s-calibration}
\begin{tabular}{lllrrrrrr}
\toprule
Dataset & Model & Seed & $\tau_s$ & $\tau_e$ & $\tau_f$ & $\epsilon_{.05}$ & $\epsilon_{.10}$ & $\epsilon_{.20}$ \\
\midrule
ECF & Qwen & 42 & -0.50 & -0.80 & -0.70 & 0.2943 & 0.5988 & 1.2278 \\
ECF & Qwen & 43 & -0.95 & -0.80 & -0.75 & 0.2553 & 0.5546 & 1.0652 \\
ECF & Qwen & 44 & -1.00 & -0.80 & -0.85 & 0.2417 & 0.5689 & 1.1142 \\
ECF & Gemma & 42 & -0.75 & -0.70 & -0.60 & 0.3254 & 0.5372 & 1.1881 \\
ECF & Gemma & 43 & -0.60 & -0.70 & -0.70 & 0.2121 & 0.5166 & 1.2271 \\
ECF & Gemma & 44 & -0.55 & -0.70 & -0.70 & 0.3025 & 0.6413 & 1.2059 \\
MEC$^4$ & Qwen & 42 & -1.20 & -1.35 & -1.20 & 0.4176 & 0.8262 & 1.6434 \\
MEC$^4$ & Qwen & 43 & -1.10 & -1.35 & -1.30 & 0.3758 & 0.6898 & 1.6585 \\
MEC$^4$ & Qwen & 44 & -1.55 & -1.35 & -1.50 & 0.3539 & 0.6963 & 1.4840 \\
MEC$^4$ & Gemma & 42 & -1.35 & -1.10 & -1.45 & 0.3825 & 0.7348 & 1.5567 \\
MEC$^4$ & Gemma & 43 & -1.20 & -1.10 & -1.25 & 0.3237 & 0.6015 & 1.3423 \\
MEC$^4$ & Gemma & 44 & -1.55 & -1.10 & -1.55 & 0.4035 & 0.6679 & 1.4610 \\
\bottomrule
\relax

\end{tabular}
\end{table*}

Table~\ref{tab:s-query-composition} expresses each student threshold as a calibrated probability and counts the accepted and rejected predictions in its nominal 10\% query band. Both sides receive substantial access without a fixed class quota. The main-text summary averages answer counts and within-student side percentages across seeds, while retaining the full threshold range. Counts use the original frozen masks, with no additional fitting.

\begin{table*}[!tbp]
\centering
\caption{Test query composition at nominal 10\% access. $p_s^*=\sigma(\tau_s)$ is the calibrated decision probability. Accepted and rejected refer to the student's predictions before consultation. Their counts sum to the actual answer count $Q$. Accepted (\%) uses $Q$ as its denominator.}
\label{tab:s-query-composition}
\begin{tabular}{@{}lllrrrrr@{}}
\toprule
Dataset & Model & Seed & $p_s^*$ & $Q$ & Accepted & Rejected & Accepted (\%) \\
\midrule
ECF & Qwen & 42 & 0.3775 & 1,544 & 675 & 869 & 43.72 \\
ECF & Qwen & 43 & 0.2789 & 1,447 & 662 & 785 & 45.75 \\
ECF & Qwen & 44 & 0.2689 & 1,480 & 689 & 791 & 46.55 \\
ECF & Gemma & 42 & 0.3208 & 1,428 & 661 & 767 & 46.29 \\
ECF & Gemma & 43 & 0.3543 & 1,395 & 709 & 686 & 50.82 \\
ECF & Gemma & 44 & 0.3659 & 1,539 & 600 & 939 & 38.99 \\
MEC$^4$ & Qwen & 42 & 0.2315 & 4,104 & 1,601 & 2,503 & 39.01 \\
MEC$^4$ & Qwen & 43 & 0.2497 & 3,877 & 1,622 & 2,255 & 41.84 \\
MEC$^4$ & Qwen & 44 & 0.1751 & 3,738 & 1,529 & 2,209 & 40.90 \\
MEC$^4$ & Gemma & 42 & 0.2059 & 4,087 & 1,631 & 2,456 & 39.91 \\
MEC$^4$ & Gemma & 43 & 0.2315 & 3,802 & 1,587 & 2,215 & 41.74 \\
MEC$^4$ & Gemma & 44 & 0.1751 & 3,826 & 1,467 & 2,359 & 38.34 \\
\bottomrule
\end{tabular}
\end{table*}

\section{Extraction Quality Across Students}
\label{app:merged-C}
\label{sec:s-results}
\subsection{Per-seed consultation gains}
\label{app:merged-C-1}
At the common nominal 10\% access condition, selective fusion improves validation and test pair F1 for all twelve students (Table~\ref{tab:s-seeds}). Replacement also improves every test baseline.

Random consultation uses the same number of answers in every dialogue but yields smaller gains. Each policy receives the expert's actual scores for its selected relations, so this difference concerns the placement of assistance.

\begin{table*}[!tbp]
\centering
\caption{Primary nominal 10\% results by independent student seed. Numbers are pair F1. Random results average 20 draws with query counts matched in each dialogue.}
\label{tab:s-seeds}
\begin{tabular}{lllrrrrrr}
\toprule
Dataset & Model & Seed & \shortstack{Val\\student} & \shortstack{Val\\selective F} & \shortstack{Test\\student} & \shortstack{Test\\random F} & \shortstack{Test\\selective F} & \shortstack{Test\\selective R} \\
\midrule
ECF & Qwen & 42 & 63.05 & 65.32 & 61.28 & 61.60 & 64.00 & 64.75 \\
ECF & Qwen & 43 & 63.12 & 65.31 & 62.04 & 62.46 & 65.11 & 65.41 \\
ECF & Qwen & 44 & 63.82 & 65.52 & 62.82 & 63.18 & 65.55 & 66.03 \\
ECF & Gemma & 42 & 61.78 & 64.52 & 61.35 & 61.84 & 64.88 & 65.18 \\
ECF & Gemma & 43 & 62.25 & 66.08 & 61.87 & 62.35 & 65.46 & 65.27 \\
ECF & Gemma & 44 & 61.63 & 65.76 & 60.56 & 61.16 & 65.09 & 65.16 \\
MEC$^4$ & Qwen & 42 & 45.18 & 49.17 & 47.74 & 48.07 & 50.27 & 50.15 \\
MEC$^4$ & Qwen & 43 & 45.30 & 47.77 & 47.10 & 47.45 & 49.87 & 49.98 \\
MEC$^4$ & Qwen & 44 & 43.59 & 47.08 & 46.46 & 46.81 & 49.09 & 49.56 \\
MEC$^4$ & Gemma & 42 & 43.44 & 46.65 & 45.65 & 46.02 & 49.18 & 49.86 \\
MEC$^4$ & Gemma & 43 & 43.24 & 47.21 & 45.14 & 45.74 & 49.80 & 49.76 \\
MEC$^4$ & Gemma & 44 & 43.26 & 46.97 & 45.71 & 46.01 & 48.10 & 49.20 \\
\bottomrule
\relax

\end{tabular}
\end{table*}

\subsection{Paired evidence for allocation}
\label{app:merged-C-2}
Every student's paired intervals favor selective fusion over both the student alone and matched random consultation (Table~\ref{tab:s-intervals}), using the dialogue-level resampling protocol in Section~\ref{sec:s-statistics}.

\begin{table*}[!tbp]
\centering
\caption{Test differences at nominal 10\% fusion. All intervals are conditional 95\% paired bootstrap intervals over dialogues. Positive values favor selective fusion.}
\label{tab:s-intervals}
\begin{tabular}{lllrcr c}
\toprule
Dataset & Model & Seed & $\Delta$ student & 95\% interval & $\Delta$ random F & 95\% interval \\
\midrule
ECF & Qwen & 42 & +2.72 & [1.61, 3.84] & +2.41 & [1.40, 3.40] \\
ECF & Qwen & 43 & +3.06 & [2.14, 3.97] & +2.65 & [1.85, 3.46] \\
ECF & Qwen & 44 & +2.73 & [1.84, 3.57] & +2.37 & [1.58, 3.08] \\
ECF & Gemma & 42 & +3.53 & [2.59, 4.45] & +3.04 & [2.19, 3.87] \\
ECF & Gemma & 43 & +3.59 & [2.64, 4.59] & +3.11 & [2.25, 4.00] \\
ECF & Gemma & 44 & +4.53 & [3.33, 5.75] & +3.93 & [2.89, 4.99] \\
MEC$^4$ & Qwen & 42 & +2.52 & [1.60, 3.46] & +2.19 & [1.36, 3.01] \\
MEC$^4$ & Qwen & 43 & +2.76 & [1.64, 3.87] & +2.41 & [1.43, 3.41] \\
MEC$^4$ & Qwen & 44 & +2.62 & [1.71, 3.47] & +2.28 & [1.47, 3.06] \\
MEC$^4$ & Gemma & 42 & +3.53 & [2.57, 4.54] & +3.16 & [2.31, 4.06] \\
MEC$^4$ & Gemma & 43 & +4.66 & [3.38, 5.91] & +4.06 & [2.91, 5.18] \\
MEC$^4$ & Gemma & 44 & +2.38 & [1.57, 3.27] & +2.09 & [1.37, 2.86] \\
\bottomrule
\relax

\end{tabular}
\end{table*}

\subsection{Emotion and cause endpoint recovery}
\label{app:merged-C-3}
Endpoint extraction projects a pair set $P$ onto the participating emotion and cause utterances within dialogue $D$:
\begin{align*}
E_D(P)&=\{i:\exists j,\ (i,j)\in P_D\},\\
C_D(P)&=\{j:\exists i,\ (i,j)\in P_D\}.
\end{align*}
Predicted projections are compared with $E_D(G)$ and $C_D(G)$ from the full gold pair set. Each dialogue--utterance identity counts once per endpoint, even if it participates in several relations. Micro precision, recall, and F1 pool these identities across dialogues. This measures recovery of emotion-bearing targets with an annotated cause and utterances used as causes. Emotion category classification and emotion utterances without an annotated pair are outside this endpoint definition.

ECF test contains 1,344 gold emotion endpoints and 1,234 cause endpoints. MEC$^4$ contains 1,849 and 1,711. The corresponding validation counts are 614/538 for ECF and 1,628/1,526 for MEC$^4$ (emotion/cause). The three-seed summary (Table~\ref{tab:s-endpoint-summary}) and per-student results (Table~\ref{tab:s-utterance-endpoints}) use the existing pair thresholds and frozen nominal 10\% masks, with no additional fitting. Both response rules improve both endpoint scores for every student. Unlike endpoint F1, pair F1 also requires correct links between those utterances.

\begin{table}[!htbp]
\centering
\caption{Emotion and cause endpoint extraction at nominal 10\% access, mean$_{\scriptscriptstyle\pm\mathrm{SD}}$ test F1 across three SFT seeds. Distinct endpoints are projected from pair outputs and full gold pairs. Bold marks the best endpoint result in each setting.}
\label{tab:s-endpoint-summary}
\begin{tabular}{@{}lllrr@{}}
\toprule
Dataset & Model & System & Emotion & Cause \\
\midrule
\multirow{3}{*}{ECF} & \multirow{3}{*}{Qwen} & Student & $79.07_{\scriptscriptstyle\pm0.60}$ & $75.52_{\scriptscriptstyle\pm0.71}$ \\
 &  & Fusion & $80.72_{\scriptscriptstyle\pm0.67}$ & $76.91_{\scriptscriptstyle\pm0.32}$ \\
 &  & Replacement & $\mathbf{80.99}_{\scriptscriptstyle\pm0.34}$ & $\mathbf{77.43}_{\scriptscriptstyle\pm0.23}$ \\
\addlinespace
\multirow{3}{*}{ECF} & \multirow{3}{*}{Gemma} & Student & $78.17_{\scriptscriptstyle\pm0.71}$ & $74.69_{\scriptscriptstyle\pm0.37}$ \\
 &  & Fusion & $80.14_{\scriptscriptstyle\pm0.65}$ & $77.41_{\scriptscriptstyle\pm0.40}$ \\
 &  & Replacement & $\mathbf{80.32}_{\scriptscriptstyle\pm0.48}$ & $\mathbf{77.41}_{\scriptscriptstyle\pm0.38}$ \\
\addlinespace
\multirow{3}{*}{MEC$^4$} & \multirow{3}{*}{Qwen} & Student & $66.78_{\scriptscriptstyle\pm0.38}$ & $59.66_{\scriptscriptstyle\pm1.01}$ \\
 &  & Fusion & $\mathbf{68.37}_{\scriptscriptstyle\pm0.35}$ & $61.74_{\scriptscriptstyle\pm0.60}$ \\
 &  & Replacement & $68.36_{\scriptscriptstyle\pm0.24}$ & $\mathbf{62.04}_{\scriptscriptstyle\pm0.49}$ \\
\addlinespace
\multirow{3}{*}{MEC$^4$} & \multirow{3}{*}{Gemma} & Student & $65.01_{\scriptscriptstyle\pm0.41}$ & $58.37_{\scriptscriptstyle\pm1.16}$ \\
 &  & Fusion & $67.96_{\scriptscriptstyle\pm0.65}$ & $\mathbf{61.12}_{\scriptscriptstyle\pm0.45}$ \\
 &  & Replacement & $\mathbf{68.50}_{\scriptscriptstyle\pm0.26}$ & $61.06_{\scriptscriptstyle\pm0.35}$ \\
\bottomrule
\end{tabular}
\end{table}

\begin{table*}[!tbp]
\centering
\caption{Test endpoint F1 for each student. F and R denote nominal 10\% selective fusion and replacement. Gold and predicted endpoints are projected from pair sets.}
\label{tab:s-utterance-endpoints}
\begin{tabular}{lllrrrrrr}
\toprule
 & & & \multicolumn{3}{c}{Emotion endpoint F1} & \multicolumn{3}{c}{Cause endpoint F1} \\
\cmidrule(lr){4-6}\cmidrule(lr){7-9}
Dataset & Model & Seed & Student & F & R & Student & F & R \\
\midrule
ECF & Qwen & 42 & 78.39 & 79.96 & 80.68 & 74.98 & 76.64 & 77.20 \\
ECF & Qwen & 43 & 79.30 & 81.01 & 80.96 & 75.25 & 76.83 & 77.44 \\
ECF & Qwen & 44 & 79.53 & 81.19 & 81.35 & 76.33 & 77.26 & 77.65 \\
\addlinespace
ECF & Gemma & 42 & 78.92 & 80.88 & 80.86 & 74.49 & 77.00 & 77.18 \\
ECF & Gemma & 43 & 78.06 & 79.71 & 80.15 & 75.11 & 77.79 & 77.85 \\
ECF & Gemma & 44 & 77.52 & 79.84 & 79.96 & 74.47 & 77.42 & 77.20 \\
\addlinespace
MEC$^4$ & Qwen & 42 & 67.21 & 68.71 & 68.51 & 60.51 & 62.30 & 62.60 \\
MEC$^4$ & Qwen & 43 & 66.49 & 68.00 & 68.47 & 58.54 & 61.11 & 61.71 \\
MEC$^4$ & Qwen & 44 & 66.65 & 68.39 & 68.08 & 59.94 & 61.81 & 61.82 \\
\addlinespace
MEC$^4$ & Gemma & 42 & 65.22 & 68.45 & 68.79 & 59.04 & 61.61 & 61.46 \\
MEC$^4$ & Gemma & 43 & 64.54 & 68.20 & 68.40 & 57.03 & 61.00 & 60.82 \\
MEC$^4$ & Gemma & 44 & 65.27 & 67.22 & 68.30 & 59.04 & 60.74 & 60.89 \\
\bottomrule
\end{tabular}
\end{table*}

\subsection{Local and distant relations}
\label{app:merged-C-4}
Table~\ref{tab:s-slices} separates self-causes from external causes. External relations are divided into the previous turn, at least two turns earlier, and future turns. Counts retain all gold in each slice, including relations outside the window. The external row pools its three component rows.

Earlier external relations also benefit from consultation (Table~\ref{tab:s-slices}), so the gains extend to more distant evidence.

Future relations reveal a limitation of the expert models on MEC$^4$. Although 22 of its 25 future gold relations are inside the window, neither independent expert recovers one. Consultation consequently remains at zero on this slice even when the required candidates are available. ECF improves under both response rules, suggesting that the difficulty also depends on the dataset.

\begin{table*}[!tbp]
\centering
\caption{Test distance slices at nominal 10\%, mean$_{\scriptscriptstyle\pm\mathrm{SD}}$ over three students. Gold counts refer to the dataset, not the sum across seeds.}
\label{tab:s-slices}
\begin{tabular}{lllrrrr}
\toprule
Dataset & Model & Slice & Gold & Student & Selective F & Selective R \\
\midrule
ECF & Qwen & self & 957 & $71.46_{\scriptscriptstyle\pm0.71}$ & $73.49_{\scriptscriptstyle\pm0.68}$ & $74.03_{\scriptscriptstyle\pm0.26}$ \\
ECF & Qwen & external & 916 & $50.46_{\scriptscriptstyle\pm3.23}$ & $54.03_{\scriptscriptstyle\pm3.45}$ & $54.55_{\scriptscriptstyle\pm2.02}$ \\
ECF & Qwen & previous turn & 534 & $60.10_{\scriptscriptstyle\pm1.55}$ & $62.73_{\scriptscriptstyle\pm1.62}$ & $62.91_{\scriptscriptstyle\pm0.80}$ \\
ECF & Qwen & earlier by at least two & 302 & $32.02_{\scriptscriptstyle\pm8.65}$ & $38.98_{\scriptscriptstyle\pm7.72}$ & $40.29_{\scriptscriptstyle\pm5.51}$ \\
ECF & Qwen & future & 80 & $24.88_{\scriptscriptstyle\pm7.68}$ & $28.83_{\scriptscriptstyle\pm6.25}$ & $29.25_{\scriptscriptstyle\pm2.71}$ \\
ECF & Gemma & self & 957 & $70.56_{\scriptscriptstyle\pm0.84}$ & $74.13_{\scriptscriptstyle\pm0.50}$ & $74.55_{\scriptscriptstyle\pm0.05}$ \\
ECF & Gemma & external & 916 & $49.48_{\scriptscriptstyle\pm0.73}$ & $53.58_{\scriptscriptstyle\pm0.58}$ & $53.24_{\scriptscriptstyle\pm0.40}$ \\
ECF & Gemma & previous turn & 534 & $61.02_{\scriptscriptstyle\pm0.29}$ & $63.96_{\scriptscriptstyle\pm0.67}$ & $63.61_{\scriptscriptstyle\pm0.27}$ \\
ECF & Gemma & earlier by at least two & 302 & $27.20_{\scriptscriptstyle\pm2.27}$ & $35.36_{\scriptscriptstyle\pm2.29}$ & $35.79_{\scriptscriptstyle\pm2.16}$ \\
ECF & Gemma & future & 80 & $14.57_{\scriptscriptstyle\pm8.51}$ & $22.11_{\scriptscriptstyle\pm8.14}$ & $20.77_{\scriptscriptstyle\pm7.17}$ \\
MEC$^4$ & Qwen & self & 1102 & $50.74_{\scriptscriptstyle\pm1.59}$ & $53.26_{\scriptscriptstyle\pm1.21}$ & $53.24_{\scriptscriptstyle\pm0.42}$ \\
MEC$^4$ & Qwen & external & 1010 & $42.16_{\scriptscriptstyle\pm0.98}$ & $45.21_{\scriptscriptstyle\pm0.25}$ & $45.66_{\scriptscriptstyle\pm0.38}$ \\
MEC$^4$ & Qwen & previous turn & 644 & $49.63_{\scriptscriptstyle\pm1.21}$ & $53.03_{\scriptscriptstyle\pm1.07}$ & $53.30_{\scriptscriptstyle\pm0.69}$ \\
MEC$^4$ & Qwen & earlier by at least two & 341 & $20.68_{\scriptscriptstyle\pm2.35}$ & $24.98_{\scriptscriptstyle\pm3.33}$ & $27.10_{\scriptscriptstyle\pm2.10}$ \\
MEC$^4$ & Qwen & future & 25 & $0.00_{\scriptscriptstyle\pm0.00}$ & $0.00_{\scriptscriptstyle\pm0.00}$ & $0.00_{\scriptscriptstyle\pm0.00}$ \\
MEC$^4$ & Gemma & self & 1102 & $48.06_{\scriptscriptstyle\pm0.30}$ & $51.62_{\scriptscriptstyle\pm1.68}$ & $52.87_{\scriptscriptstyle\pm0.86}$ \\
MEC$^4$ & Gemma & external & 1010 & $41.91_{\scriptscriptstyle\pm0.13}$ & $45.44_{\scriptscriptstyle\pm0.22}$ & $45.12_{\scriptscriptstyle\pm0.31}$ \\
MEC$^4$ & Gemma & previous turn & 644 & $48.55_{\scriptscriptstyle\pm0.59}$ & $52.64_{\scriptscriptstyle\pm0.19}$ & $52.85_{\scriptscriptstyle\pm0.08}$ \\
MEC$^4$ & Gemma & earlier by at least two & 341 & $23.74_{\scriptscriptstyle\pm0.86}$ & $25.55_{\scriptscriptstyle\pm0.48}$ & $24.28_{\scriptscriptstyle\pm0.74}$ \\
MEC$^4$ & Gemma & future & 25 & $0.00_{\scriptscriptstyle\pm0.00}$ & $0.00_{\scriptscriptstyle\pm0.00}$ & $0.00_{\scriptscriptstyle\pm0.00}$ \\
\bottomrule
\relax

\end{tabular}
\end{table*}

\section{Granularity and Complete Cause Sets}
\label{app:merged-D}
\label{sec:s-granularity}
\subsection{The opportunity cost of grouped answers}
\label{app:merged-D-1}
Target actions request all candidate answers for one target. Dialogue actions request all candidates in a conversation. Every granularity uses the same pairwise expert interface. Group priority is either minimum boundary distance or mean $\sigma(-|z_s-\tau_s|)$, selected on validation for each budget. A group is admitted only if it fits the remaining global allowance. Oversized groups are skipped, leaving later groups eligible.

Relation selection retains its advantage when matched to the exact number of answers used by each group policy (Table~\ref{tab:s-granularity-exact}). All paired intervals are positive. Requiring a whole group to reach a useful relation lowers the quality of the selected set even at the same actual access.

\begin{table*}[!tbp]
\centering
\caption{Actual group access and the relation comparator at exactly that access, under fixed fusion. $\Delta$: matched Relation minus Group.}
\label{tab:s-granularity-exact}
\begin{tabular}{@{}llrlrrrr@{}}
\toprule
Dataset & Model & $b$ (\%) & Group & $Q$ & Group F1 & Relation F1 & $\Delta$ [95\% CI] \\
\midrule
ECF & Qwen & 5 & Target & 800.0 & $62.57_{\scriptscriptstyle\pm0.54}$ & $63.98_{\scriptscriptstyle\pm0.71}$ & +1.41 [+1.00, +1.79] \\
ECF & Qwen & 5 & Dialogue & 799.3 & $62.32_{\scriptscriptstyle\pm0.85}$ & $63.98_{\scriptscriptstyle\pm0.71}$ & +1.66 [+1.12, +2.16] \\
ECF & Qwen & 10 & Target & 1601.0 & $63.08_{\scriptscriptstyle\pm0.63}$ & $64.95_{\scriptscriptstyle\pm0.81}$ & +1.88 [+1.37, +2.36] \\
ECF & Qwen & 10 & Dialogue & 1599.7 & $62.70_{\scriptscriptstyle\pm1.02}$ & $64.95_{\scriptscriptstyle\pm0.82}$ & +2.25 [+1.59, +2.89] \\
ECF & Qwen & 20 & Target & 3203.0 & $63.61_{\scriptscriptstyle\pm0.92}$ & $65.51_{\scriptscriptstyle\pm0.80}$ & +1.90 [+1.36, +2.42] \\
ECF & Qwen & 20 & Dialogue & 3200.3 & $63.28_{\scriptscriptstyle\pm1.05}$ & $65.51_{\scriptscriptstyle\pm0.80}$ & +2.24 [+1.58, +2.90] \\
ECF & Gemma & 5 & Target & 800.0 & $62.13_{\scriptscriptstyle\pm0.47}$ & $64.06_{\scriptscriptstyle\pm0.19}$ & +1.93 [+1.38, +2.47] \\
ECF & Gemma & 5 & Dialogue & 799.3 & $61.54_{\scriptscriptstyle\pm0.77}$ & $64.06_{\scriptscriptstyle\pm0.19}$ & +2.52 [+1.94, +3.07] \\
ECF & Gemma & 10 & Target & 1601.0 & $62.71_{\scriptscriptstyle\pm0.50}$ & $65.22_{\scriptscriptstyle\pm0.47}$ & +2.51 [+1.89, +3.14] \\
ECF & Gemma & 10 & Dialogue & 1600.7 & $61.86_{\scriptscriptstyle\pm0.93}$ & $65.22_{\scriptscriptstyle\pm0.47}$ & +3.35 [+2.65, +4.09] \\
ECF & Gemma & 20 & Target & 3203.0 & $63.75_{\scriptscriptstyle\pm0.65}$ & $65.99_{\scriptscriptstyle\pm0.35}$ & +2.24 [+1.67, +2.82] \\
ECF & Gemma & 20 & Dialogue & 3203.0 & $62.38_{\scriptscriptstyle\pm0.66}$ & $65.99_{\scriptscriptstyle\pm0.35}$ & +3.61 [+2.86, +4.39] \\
MEC$^4$ & Qwen & 5 & Target & 2022.7 & $47.56_{\scriptscriptstyle\pm0.69}$ & $49.15_{\scriptscriptstyle\pm0.77}$ & +1.58 [+1.11, +2.02] \\
MEC$^4$ & Qwen & 5 & Dialogue & 2022.7 & $47.35_{\scriptscriptstyle\pm0.61}$ & $49.15_{\scriptscriptstyle\pm0.77}$ & +1.80 [+1.24, +2.36] \\
MEC$^4$ & Qwen & 10 & Target & 4046.0 & $47.97_{\scriptscriptstyle\pm0.60}$ & $49.77_{\scriptscriptstyle\pm0.59}$ & +1.80 [+1.24, +2.33] \\
MEC$^4$ & Qwen & 10 & Dialogue & 4044.7 & $47.51_{\scriptscriptstyle\pm0.60}$ & $49.77_{\scriptscriptstyle\pm0.59}$ & +2.25 [+1.60, +2.89] \\
MEC$^4$ & Qwen & 20 & Target & 8091.3 & $48.56_{\scriptscriptstyle\pm0.57}$ & $49.90_{\scriptscriptstyle\pm0.54}$ & +1.34 [+0.90, +1.78] \\
MEC$^4$ & Qwen & 20 & Dialogue & 8089.7 & $48.09_{\scriptscriptstyle\pm0.62}$ & $49.90_{\scriptscriptstyle\pm0.54}$ & +1.81 [+1.18, +2.44] \\
MEC$^4$ & Gemma & 5 & Target & 2023.0 & $45.97_{\scriptscriptstyle\pm0.14}$ & $48.18_{\scriptscriptstyle\pm0.34}$ & +2.21 [+1.71, +2.70] \\
MEC$^4$ & Gemma & 5 & Dialogue & 2017.3 & $45.57_{\scriptscriptstyle\pm0.37}$ & $48.16_{\scriptscriptstyle\pm0.32}$ & +2.59 [+1.99, +3.20] \\
MEC$^4$ & Gemma & 10 & Target & 4046.0 & $46.55_{\scriptscriptstyle\pm0.44}$ & $49.06_{\scriptscriptstyle\pm0.87}$ & +2.51 [+1.93, +3.08] \\
MEC$^4$ & Gemma & 10 & Dialogue & 4042.0 & $45.74_{\scriptscriptstyle\pm0.28}$ & $49.06_{\scriptscriptstyle\pm0.87}$ & +3.31 [+2.54, +4.07] \\
MEC$^4$ & Gemma & 20 & Target & 8092.0 & $47.46_{\scriptscriptstyle\pm0.58}$ & $49.45_{\scriptscriptstyle\pm1.01}$ & +1.99 [+1.49, +2.46] \\
MEC$^4$ & Gemma & 20 & Dialogue & 8086.0 & $46.10_{\scriptscriptstyle\pm0.16}$ & $49.45_{\scriptscriptstyle\pm1.01}$ & +3.35 [+2.58, +4.08] \\
\bottomrule
\end{tabular}
\end{table*}

\subsection{Recovery of an entire cause set}
\label{app:merged-D-2}
Complete cause sets test whether better relation allocation also recovers all the annotated causes for a target (Table~\ref{tab:s-cause-sets}). Success requires exact equality between predicted and annotated sets, including gold outside the window. The subset with multiple causes includes targets with at least two annotated causes. The broader nonempty subset includes targets with one or more, excluding targets without an annotated cause from both denominators.

Relation selection has the highest recovery of complete sets throughout Table~\ref{tab:s-cause-sets}. Directing help to individual relations is more effective here than spending the same budget on all candidates for a target.

\begin{table*}[!tbp]
\centering
\caption{Exact recovery of cause sets (\%) at a hard 10\% answer cap. $n$ is the number of eligible gold targets, counted once per dataset. Row maxima are bold.}
\label{tab:s-cause-sets}
\begin{tabular}{@{}lllrrrr@{}}
\toprule
Dataset & Model & Gold target subset & $n$ & Relation & Target & Dialogue \\
\midrule
ECF & Qwen & At least one cause & 1344 & $\mathbf{48.26}_{\scriptscriptstyle\pm0.55}$ & $45.86_{\scriptscriptstyle\pm0.71}$ & $45.56_{\scriptscriptstyle\pm0.62}$ \\
ECF & Qwen & At least two causes & 417 & $\mathbf{38.85}_{\scriptscriptstyle\pm6.17}$ & $35.73_{\scriptscriptstyle\pm4.99}$ & $35.09_{\scriptscriptstyle\pm5.48}$ \\
ECF & Gemma & At least one cause & 1344 & $\mathbf{48.29}_{\scriptscriptstyle\pm0.65}$ & $45.16_{\scriptscriptstyle\pm0.76}$ & $44.64_{\scriptscriptstyle\pm0.84}$ \\
ECF & Gemma & At least two causes & 417 & $\mathbf{34.69}_{\scriptscriptstyle\pm2.45}$ & $31.65_{\scriptscriptstyle\pm2.59}$ & $31.25_{\scriptscriptstyle\pm2.56}$ \\
MEC$^4$ & Qwen & At least one cause & 1849 & $\mathbf{37.79}_{\scriptscriptstyle\pm0.19}$ & $34.36_{\scriptscriptstyle\pm1.68}$ & $33.80_{\scriptscriptstyle\pm2.08}$ \\
MEC$^4$ & Qwen & At least two causes & 218 & $\mathbf{18.81}_{\scriptscriptstyle\pm2.75}$ & $17.28_{\scriptscriptstyle\pm1.74}$ & $16.67_{\scriptscriptstyle\pm2.26}$ \\
MEC$^4$ & Gemma & At least one cause & 1849 & $\mathbf{37.28}_{\scriptscriptstyle\pm0.95}$ & $32.86_{\scriptscriptstyle\pm3.84}$ & $31.89_{\scriptscriptstyle\pm4.79}$ \\
MEC$^4$ & Gemma & At least two causes & 218 & $\mathbf{17.28}_{\scriptscriptstyle\pm3.74}$ & $14.53_{\scriptscriptstyle\pm1.61}$ & $13.91_{\scriptscriptstyle\pm1.91}$ \\
\bottomrule
\end{tabular}
\end{table*}

\section{Competing Query Rankings}
\label{app:merged-E}
\label{sec:s-router}
\subsection{Inputs to the learned selector}
\label{app:merged-E-1}
Table~\ref{tab:s-features} defines the 16 features available to the student and supplied to the learned regressor. Candidate statistics use all relations for the same target. Scores are calibrated student logits, text lengths count Python Unicode characters, and standard deviations use population normalization. Selection features exclude expert scores and gold emotion labels.

Beyond confidence, these features could distinguish an isolated borderline relation from a target with several plausible causes.

\begin{table*}[!tbp]
\centering
\caption{The 16 features available to the learned revision value regressor. $\mathcal C_i$ contains the target's candidate relations and $n$ is dialogue length.}
\label{tab:s-features}
\begin{tabular}{@{}rlp{11.8cm}@{}}
\toprule
No. & Feature & Definition \\
\midrule
1 & Student score & Calibrated logit $z_s(i,j)$. \\
2 & Boundary distance & $|z_s(i,j)-\tau_s|$. \\
3 & Student class & $\mathbb 1[z_s(i,j)>\tau_s]$. \\
4 & Signed distance & $j-i$. Positive values denote later candidate utterances. \\
5 & Speaker match & Equality of target and candidate speaker strings. \\
6 & Candidate length & Character count of candidate utterance text. \\
7 & Target length & Character count of target utterance text. \\
8 & Target position & $i/n$, using one-based turn indices. \\
9 & Dialogue length & Number of utterances $n$. \\
10 & Candidate rank & Number of relations in $\mathcal C_i$ with a strictly larger student logit. \\
11 & Target maximum & Maximum student logit over $\mathcal C_i$. \\
12 & Target mean & Mean student logit over $\mathcal C_i$. \\
13 & Target dispersion & Population standard deviation of student logits over $\mathcal C_i$. \\
14 & Positive count & Number of student-positive relations in $\mathcal C_i$. \\
15 & Candidate count & $|\mathcal C_i|$. \\
16 & Self relation & $\mathbb 1[i=j]$. \\
\bottomrule
\end{tabular}
\end{table*}

\subsection{Learning objectives and fitting}
\label{app:merged-E-2}
The router uses histogram gradient boosting with squared-error loss, at most 255 bins per feature, 120 iterations, at most seven leaves per tree, minimum leaf size 40, learning rate 0.05, $L_2$ regularization 1, random seed 3417, and no early stopping. Five GroupKFold partitions separate dialogues. Out-of-fold validation F1 selects between two training targets under the relevant response rule, access constraint, and budget. A tie selects count improvement. The final model is refitted on all validation dialogues. Student and expert score calibrations remain the common frozen validation maps throughout.

For binary gold $y_r$, student decision $b_r$, and decision $h_r$ after querying, define $\Delta_r=h_r-b_r$. Let $f$ be the student's F1 in $[0,1]$ on the fitting dialogues, retaining all their gold pairs. The target for count improvement is
\begin{equation}
u_r^{\mathrm{count}}=(2y_r-f)\Delta_r.
\end{equation}
This target rewards recovered positives and removed false positives and penalizes the reverse changes. Summing it over a set of revisions gives the sign of the F1 change relative to the baseline on the fitting dialogues. Its magnitude is a training utility rather than the F1 difference itself.

The alternative target is reduction in logistic loss. With $v_r=2y_r-1$ and score $z_h$ after querying, it is
\begin{equation}
u_r^{\mathrm{loss}}=\log(1+e^{-v_rz_s})-\log(1+e^{-v_rz_h}).
\end{equation}
Fusion uses $z_h=(z_s+z_e)/2$ and threshold $\tau_s$. Replacement uses $z_h=z_e$ and threshold $\tau_e$. Both targets are rebuilt for replacement, so each router learns the consequences of its own response rule. Expert outputs construct fitting targets only. During selection, the regressor receives the student features above.

Native entropy ranks the binary distribution induced by the raw yes/no margin. Calibrated entropy ranks $\sigma(z_s)$. They rank relations by increasing absolute raw margin and increasing $|z_s|$, respectively. Boundary priority is increasing $|z_s-\tau_s|$. The comparison therefore distinguishes a center at probability $0.5$ from the F1 decision threshold. A nonzero affine score transform preserves distance ranking around its corresponding transformed threshold.

Count improvement is selected most often, but reduction in logistic loss is preferred for some students, especially at larger allowances (Table~\ref{tab:s-router-objectives}). The preferred target depends on the query budget and response rule.

\begin{table}[!htbp]
\centering
\caption{Selected router targets at 5/10/20\%. C denotes count improvement and L reduction in logistic loss. F and R denote fusion and replacement. M matches dialogue counts, and G uses a global hard budget.}
\label{tab:s-router-objectives}
\begin{tabular}{@{}llrrrrr@{}}
\toprule
Dataset & Model & Seed & F-M & F-G & R-M & R-G \\
\midrule
ECF & Qwen & 42 & C/C/C & C/C/C & L/C/L & L/C/L \\
ECF & Qwen & 43 & C/C/C & C/C/C & C/C/C & C/C/L \\
ECF & Qwen & 44 & C/C/C & C/C/C & C/C/C & C/C/C \\
ECF & Gemma & 42 & C/C/L & C/C/C & C/C/C & C/C/C \\
ECF & Gemma & 43 & C/C/C & C/C/C & C/C/C & C/C/C \\
ECF & Gemma & 44 & C/C/C & C/C/C & C/C/C & C/C/C \\
MEC$^4$ & Qwen & 42 & C/C/L & C/C/L & C/C/L & C/C/L \\
MEC$^4$ & Qwen & 43 & L/L/L & L/L/L & C/C/L & C/C/L \\
MEC$^4$ & Qwen & 44 & C/C/L & C/C/L & C/C/C & C/C/C \\
MEC$^4$ & Gemma & 42 & C/C/L & C/C/L & L/C/L & L/C/C \\
MEC$^4$ & Gemma & 43 & C/C/L & C/C/L & C/C/C & C/C/L \\
MEC$^4$ & Gemma & 44 & C/C/C & C/C/L & C/L/L & C/L/C \\
\bottomrule
\end{tabular}
\end{table}

\subsection{Fusion within and across dialogues}
\label{app:merged-E-3}
\label{sec:s-crossed}
The comparison within dialogues gives every selector the boundary policy's exact query count in each dialogue. The global comparison gives each selector $K=\lfloor bN\rfloor$ answers over the full candidate pool and lets it distribute those answers independently. Ties retain lexicographic dialogue--target--candidate order. Both use identical models, score maps fitted on validation, and response rules.

With a fixed allowance in each dialogue, Boundary has the highest mean at the primary budget in every setting (Table~\ref{tab:s-cross-fusion-dialogue_matched}). At wider access on MEC$^4$, entropy nearly catches up, as admitting more candidates reduces the effect of differences in their ranking.

Allowing each policy to distribute its budget across conversations preserves Boundary's mean ordering at the primary budget (Table~\ref{tab:s-cross-fusion-global_hard}). In both comparisons, some paired intervals include zero, leaving the advantage uncertain in those settings.

\begin{table*}[!tbp]
\centering
\caption{Fusion with matched query counts per dialogue. $Q$: mean answers. $\Delta$: Boundary minus Learned with a 95\% paired interval.}
\label{tab:s-cross-fusion-dialogue_matched}
\begin{tabular}{@{}llrrrrrrr@{}}
\toprule
Dataset & Model & $b$ (\%) & $Q$ & $H_\ell$ & $H_p$ & Learned & Boundary & $\Delta$ [95\% CI] \\
\midrule
ECF & Qwen & 5 & 661.3 & $62.81_{\scriptscriptstyle\pm0.76}$ & $62.44_{\scriptscriptstyle\pm0.80}$ & $63.63_{\scriptscriptstyle\pm0.44}$ & $\mathbf{63.69}_{\scriptscriptstyle\pm0.63}$ & +0.06 [-0.28, +0.40] \\
ECF & Qwen & 10 & 1490.3 & $64.09_{\scriptscriptstyle\pm0.59}$ & $63.58_{\scriptscriptstyle\pm0.60}$ & $64.47_{\scriptscriptstyle\pm0.50}$ & $\mathbf{64.89}_{\scriptscriptstyle\pm0.79}$ & +0.42 [+0.04, +0.77] \\
ECF & Qwen & 20 & 3108.3 & $65.36_{\scriptscriptstyle\pm0.70}$ & $65.32_{\scriptscriptstyle\pm0.72}$ & $64.96_{\scriptscriptstyle\pm0.58}$ & $\mathbf{65.51}_{\scriptscriptstyle\pm0.80}$ & +0.55 [+0.20, +0.89] \\
ECF & Gemma & 5 & 714.0 & $62.62_{\scriptscriptstyle\pm1.32}$ & $62.25_{\scriptscriptstyle\pm0.51}$ & $63.59_{\scriptscriptstyle\pm0.28}$ & $\mathbf{63.95}_{\scriptscriptstyle\pm0.20}$ & +0.36 [+0.00, +0.73] \\
ECF & Gemma & 10 & 1454.0 & $64.69_{\scriptscriptstyle\pm0.67}$ & $64.18_{\scriptscriptstyle\pm0.38}$ & $64.68_{\scriptscriptstyle\pm0.50}$ & $\mathbf{65.14}_{\scriptscriptstyle\pm0.29}$ & +0.47 [+0.14, +0.81] \\
ECF & Gemma & 20 & 3094.0 & $65.89_{\scriptscriptstyle\pm0.45}$ & $65.84_{\scriptscriptstyle\pm0.40}$ & $65.01_{\scriptscriptstyle\pm0.71}$ & $\mathbf{65.99}_{\scriptscriptstyle\pm0.36}$ & +0.98 [+0.57, +1.40] \\
MEC$^4$ & Qwen & 5 & 2004.7 & $47.93_{\scriptscriptstyle\pm0.80}$ & $47.85_{\scriptscriptstyle\pm0.78}$ & $48.72_{\scriptscriptstyle\pm0.76}$ & $\mathbf{49.10}_{\scriptscriptstyle\pm0.74}$ & +0.38 [+0.01, +0.75] \\
MEC$^4$ & Qwen & 10 & 3906.3 & $49.40_{\scriptscriptstyle\pm0.96}$ & $49.41_{\scriptscriptstyle\pm0.96}$ & $49.47_{\scriptscriptstyle\pm0.45}$ & $\mathbf{49.74}_{\scriptscriptstyle\pm0.60}$ & +0.27 [-0.02, +0.56] \\
MEC$^4$ & Qwen & 20 & 8314.3 & $\mathbf{49.91}_{\scriptscriptstyle\pm0.53}$ & $\mathbf{49.91}_{\scriptscriptstyle\pm0.53}$ & $49.66_{\scriptscriptstyle\pm0.61}$ & $49.90_{\scriptscriptstyle\pm0.54}$ & +0.24 [+0.06, +0.42] \\
MEC$^4$ & Gemma & 5 & 1854.7 & $46.43_{\scriptscriptstyle\pm0.12}$ & $46.44_{\scriptscriptstyle\pm0.16}$ & $47.52_{\scriptscriptstyle\pm0.40}$ & $\mathbf{48.08}_{\scriptscriptstyle\pm0.20}$ & +0.56 [+0.24, +0.88] \\
MEC$^4$ & Gemma & 10 & 3905.0 & $48.64_{\scriptscriptstyle\pm0.92}$ & $48.59_{\scriptscriptstyle\pm0.82}$ & $48.56_{\scriptscriptstyle\pm0.79}$ & $\mathbf{49.02}_{\scriptscriptstyle\pm0.86}$ & +0.46 [+0.21, +0.72] \\
MEC$^4$ & Gemma & 20 & 8141.3 & $49.45_{\scriptscriptstyle\pm1.01}$ & $49.45_{\scriptscriptstyle\pm1.01}$ & $49.16_{\scriptscriptstyle\pm1.00}$ & $\mathbf{49.45}_{\scriptscriptstyle\pm1.01}$ & +0.29 [+0.07, +0.50] \\
\bottomrule
\end{tabular}
\end{table*}
\begin{table*}[!tbp]
\centering
\caption{Fusion under a global hard budget. $Q$: mean answers. $\Delta$: Boundary minus Learned with a 95\% paired interval.}
\label{tab:s-cross-fusion-global_hard}
\begin{tabular}{@{}llrrrrrrr@{}}
\toprule
Dataset & Model & $b$ (\%) & $Q$ & $H_\ell$ & $H_p$ & Learned & Boundary & $\Delta$ [95\% CI] \\
\midrule
ECF & Qwen & 5 & 800.0 & $62.87_{\scriptscriptstyle\pm0.62}$ & $62.43_{\scriptscriptstyle\pm0.61}$ & $63.77_{\scriptscriptstyle\pm0.49}$ & $\mathbf{63.98}_{\scriptscriptstyle\pm0.71}$ & +0.21 [-0.17, +0.57] \\
ECF & Qwen & 10 & 1601.0 & $64.12_{\scriptscriptstyle\pm0.69}$ & $63.71_{\scriptscriptstyle\pm0.64}$ & $64.52_{\scriptscriptstyle\pm0.62}$ & $\mathbf{64.95}_{\scriptscriptstyle\pm0.81}$ & +0.44 [+0.04, +0.80] \\
ECF & Qwen & 20 & 3203.0 & $65.47_{\scriptscriptstyle\pm0.80}$ & $65.44_{\scriptscriptstyle\pm0.81}$ & $64.95_{\scriptscriptstyle\pm0.55}$ & $\mathbf{65.51}_{\scriptscriptstyle\pm0.80}$ & +0.56 [+0.20, +0.90] \\
ECF & Gemma & 5 & 800.0 & $62.69_{\scriptscriptstyle\pm1.57}$ & $61.95_{\scriptscriptstyle\pm0.78}$ & $63.78_{\scriptscriptstyle\pm0.42}$ & $\mathbf{64.06}_{\scriptscriptstyle\pm0.19}$ & +0.28 [-0.13, +0.72] \\
ECF & Gemma & 10 & 1601.0 & $64.77_{\scriptscriptstyle\pm0.80}$ & $64.32_{\scriptscriptstyle\pm0.70}$ & $64.91_{\scriptscriptstyle\pm0.60}$ & $\mathbf{65.22}_{\scriptscriptstyle\pm0.47}$ & +0.31 [-0.03, +0.67] \\
ECF & Gemma & 20 & 3203.0 & $65.93_{\scriptscriptstyle\pm0.42}$ & $65.92_{\scriptscriptstyle\pm0.47}$ & $65.20_{\scriptscriptstyle\pm0.51}$ & $\mathbf{65.99}_{\scriptscriptstyle\pm0.35}$ & +0.79 [+0.38, +1.22] \\
MEC$^4$ & Qwen & 5 & 2023.0 & $47.48_{\scriptscriptstyle\pm0.78}$ & $47.45_{\scriptscriptstyle\pm0.76}$ & $48.67_{\scriptscriptstyle\pm0.88}$ & $\mathbf{49.14}_{\scriptscriptstyle\pm0.77}$ & +0.46 [+0.11, +0.83] \\
MEC$^4$ & Qwen & 10 & 4046.0 & $49.54_{\scriptscriptstyle\pm0.79}$ & $49.54_{\scriptscriptstyle\pm0.79}$ & $49.48_{\scriptscriptstyle\pm0.41}$ & $\mathbf{49.77}_{\scriptscriptstyle\pm0.59}$ & +0.29 [-0.01, +0.59] \\
MEC$^4$ & Qwen & 20 & 8092.0 & $49.89_{\scriptscriptstyle\pm0.56}$ & $49.89_{\scriptscriptstyle\pm0.56}$ & $49.71_{\scriptscriptstyle\pm0.65}$ & $\mathbf{49.90}_{\scriptscriptstyle\pm0.54}$ & +0.20 [+0.03, +0.36] \\
MEC$^4$ & Gemma & 5 & 2023.0 & $46.12_{\scriptscriptstyle\pm0.25}$ & $46.09_{\scriptscriptstyle\pm0.21}$ & $47.80_{\scriptscriptstyle\pm0.42}$ & $\mathbf{48.18}_{\scriptscriptstyle\pm0.34}$ & +0.38 [+0.02, +0.74] \\
MEC$^4$ & Gemma & 10 & 4046.0 & $48.80_{\scriptscriptstyle\pm0.84}$ & $48.79_{\scriptscriptstyle\pm0.83}$ & $48.69_{\scriptscriptstyle\pm0.78}$ & $\mathbf{49.06}_{\scriptscriptstyle\pm0.87}$ & +0.37 [+0.13, +0.62] \\
MEC$^4$ & Gemma & 20 & 8092.0 & $49.44_{\scriptscriptstyle\pm1.02}$ & $49.44_{\scriptscriptstyle\pm1.02}$ & $49.32_{\scriptscriptstyle\pm0.85}$ & $\mathbf{49.45}_{\scriptscriptstyle\pm1.01}$ & +0.13 [-0.06, +0.33] \\
\bottomrule
\end{tabular}
\end{table*}

\subsection{Ranking with direct replacement}
\label{app:merged-E-4}
Boundary also retains the higher mean at the primary budget when expert decisions replace student predictions directly (Tables~\ref{tab:s-cross-replacement-dialogue_matched} and~\ref{tab:s-cross-replacement-global_hard}). Each learned router is fitted for replacement, including objective selection. The comparison therefore tests selection under a rule that can overturn confident student judgments without averaging.

Under global access, both Qwen settings have positive 10\% paired intervals, whereas the Gemma intervals include zero. At 20\%, native entropy is slightly higher on Qwen/ECF, and the MEC$^4$ selectors are close. Together with the fusion results, this pattern shows that boundary ranking remains competitive as the response rule changes, with its advantage depending on budget and model pairing.

\begin{table*}[!tbp]
\centering
\caption{Replacement with matched query counts per dialogue. $Q$: mean answers. $\Delta$: Boundary minus Learned with a 95\% paired interval.}
\label{tab:s-cross-replacement-dialogue_matched}
\begin{tabular}{@{}llrrrrrrr@{}}
\toprule
Dataset & Model & $b$ (\%) & $Q$ & $H_\ell$ & $H_p$ & Learned & Boundary & $\Delta$ [95\% CI] \\
\midrule
ECF & Qwen & 5 & 661.3 & $62.77_{\scriptscriptstyle\pm0.64}$ & $62.46_{\scriptscriptstyle\pm0.64}$ & $63.70_{\scriptscriptstyle\pm0.99}$ & $\mathbf{63.92}_{\scriptscriptstyle\pm0.58}$ & +0.22 [-0.24, +0.66] \\
ECF & Qwen & 10 & 1490.3 & $64.47_{\scriptscriptstyle\pm0.48}$ & $64.00_{\scriptscriptstyle\pm0.23}$ & $64.69_{\scriptscriptstyle\pm0.36}$ & $\mathbf{65.40}_{\scriptscriptstyle\pm0.64}$ & +0.71 [+0.29, +1.11] \\
ECF & Qwen & 20 & 3108.3 & $66.30_{\scriptscriptstyle\pm0.13}$ & $66.21_{\scriptscriptstyle\pm0.15}$ & $65.12_{\scriptscriptstyle\pm0.41}$ & $\mathbf{66.33}_{\scriptscriptstyle\pm0.05}$ & +1.21 [+0.60, +1.77] \\
ECF & Gemma & 5 & 714.0 & $62.46_{\scriptscriptstyle\pm1.44}$ & $61.96_{\scriptscriptstyle\pm0.48}$ & $63.56_{\scriptscriptstyle\pm0.48}$ & $\mathbf{63.91}_{\scriptscriptstyle\pm0.20}$ & +0.35 [-0.08, +0.79] \\
ECF & Gemma & 10 & 1454.0 & $64.56_{\scriptscriptstyle\pm0.62}$ & $64.04_{\scriptscriptstyle\pm0.17}$ & $64.80_{\scriptscriptstyle\pm0.59}$ & $\mathbf{65.21}_{\scriptscriptstyle\pm0.05}$ & +0.40 [-0.13, +0.93] \\
ECF & Gemma & 20 & 3094.0 & $66.27_{\scriptscriptstyle\pm0.32}$ & $66.13_{\scriptscriptstyle\pm0.23}$ & $65.68_{\scriptscriptstyle\pm0.45}$ & $\mathbf{66.43}_{\scriptscriptstyle\pm0.19}$ & +0.75 [+0.26, +1.26] \\
MEC$^4$ & Qwen & 5 & 2004.7 & $47.94_{\scriptscriptstyle\pm0.64}$ & $47.88_{\scriptscriptstyle\pm0.59}$ & $48.85_{\scriptscriptstyle\pm0.13}$ & $\mathbf{49.14}_{\scriptscriptstyle\pm0.48}$ & +0.29 [-0.18, +0.77] \\
MEC$^4$ & Qwen & 10 & 3906.3 & $49.79_{\scriptscriptstyle\pm0.78}$ & $49.80_{\scriptscriptstyle\pm0.78}$ & $49.58_{\scriptscriptstyle\pm0.26}$ & $\mathbf{49.90}_{\scriptscriptstyle\pm0.31}$ & +0.32 [-0.05, +0.68] \\
MEC$^4$ & Qwen & 20 & 8314.3 & $\mathbf{49.86}_{\scriptscriptstyle\pm0.08}$ & $\mathbf{49.86}_{\scriptscriptstyle\pm0.08}$ & $49.65_{\scriptscriptstyle\pm0.10}$ & $49.86_{\scriptscriptstyle\pm0.09}$ & +0.21 [-0.16, +0.58] \\
MEC$^4$ & Gemma & 5 & 1854.7 & $46.30_{\scriptscriptstyle\pm0.39}$ & $46.31_{\scriptscriptstyle\pm0.56}$ & $47.93_{\scriptscriptstyle\pm0.63}$ & $\mathbf{48.32}_{\scriptscriptstyle\pm0.30}$ & +0.39 [-0.15, +0.93] \\
MEC$^4$ & Gemma & 10 & 3905.0 & $49.00_{\scriptscriptstyle\pm0.21}$ & $48.97_{\scriptscriptstyle\pm0.34}$ & $49.19_{\scriptscriptstyle\pm0.49}$ & $\mathbf{49.60}_{\scriptscriptstyle\pm0.36}$ & +0.42 [-0.11, +0.90] \\
MEC$^4$ & Gemma & 20 & 8141.3 & $\mathbf{50.26}_{\scriptscriptstyle\pm0.07}$ & $\mathbf{50.26}_{\scriptscriptstyle\pm0.07}$ & $49.90_{\scriptscriptstyle\pm0.25}$ & $50.23_{\scriptscriptstyle\pm0.04}$ & +0.33 [-0.14, +0.81] \\
\bottomrule
\end{tabular}
\end{table*}
\begin{table*}[!tbp]
\centering
\caption{Replacement under a global hard budget. $Q$: mean answers. $\Delta$: Boundary minus Learned with a 95\% paired interval.}
\label{tab:s-cross-replacement-global_hard}
\begin{tabular}{@{}llrrrrrrr@{}}
\toprule
Dataset & Model & $b$ (\%) & $Q$ & $H_\ell$ & $H_p$ & Learned & Boundary & $\Delta$ [95\% CI] \\
\midrule
ECF & Qwen & 5 & 800.0 & $63.00_{\scriptscriptstyle\pm0.69}$ & $62.52_{\scriptscriptstyle\pm0.58}$ & $63.63_{\scriptscriptstyle\pm1.12}$ & $\mathbf{64.22}_{\scriptscriptstyle\pm0.70}$ & +0.59 [+0.08, +1.12] \\
ECF & Qwen & 10 & 1601.0 & $64.44_{\scriptscriptstyle\pm0.68}$ & $64.10_{\scriptscriptstyle\pm0.55}$ & $64.87_{\scriptscriptstyle\pm0.39}$ & $\mathbf{65.45}_{\scriptscriptstyle\pm0.65}$ & +0.58 [+0.14, +1.01] \\
ECF & Qwen & 20 & 3203.0 & $\mathbf{66.38}_{\scriptscriptstyle\pm0.21}$ & $66.24_{\scriptscriptstyle\pm0.22}$ & $65.21_{\scriptscriptstyle\pm0.59}$ & $66.32_{\scriptscriptstyle\pm0.07}$ & +1.11 [+0.48, +1.69] \\
ECF & Gemma & 5 & 800.0 & $62.58_{\scriptscriptstyle\pm1.68}$ & $61.66_{\scriptscriptstyle\pm0.70}$ & $63.84_{\scriptscriptstyle\pm0.50}$ & $\mathbf{64.01}_{\scriptscriptstyle\pm0.16}$ & +0.17 [-0.29, +0.67] \\
ECF & Gemma & 10 & 1601.0 & $64.70_{\scriptscriptstyle\pm0.85}$ & $64.20_{\scriptscriptstyle\pm0.46}$ & $64.98_{\scriptscriptstyle\pm0.65}$ & $\mathbf{65.28}_{\scriptscriptstyle\pm0.39}$ & +0.29 [-0.25, +0.84] \\
ECF & Gemma & 20 & 3203.0 & $66.37_{\scriptscriptstyle\pm0.38}$ & $66.35_{\scriptscriptstyle\pm0.29}$ & $65.63_{\scriptscriptstyle\pm0.39}$ & $\mathbf{66.44}_{\scriptscriptstyle\pm0.21}$ & +0.81 [+0.35, +1.32] \\
MEC$^4$ & Qwen & 5 & 2023.0 & $47.43_{\scriptscriptstyle\pm0.48}$ & $47.40_{\scriptscriptstyle\pm0.48}$ & $48.80_{\scriptscriptstyle\pm0.23}$ & $\mathbf{49.17}_{\scriptscriptstyle\pm0.52}$ & +0.37 [-0.15, +0.91] \\
MEC$^4$ & Qwen & 10 & 4046.0 & $49.74_{\scriptscriptstyle\pm0.18}$ & $49.74_{\scriptscriptstyle\pm0.17}$ & $49.57_{\scriptscriptstyle\pm0.14}$ & $\mathbf{49.98}_{\scriptscriptstyle\pm0.31}$ & +0.42 [+0.03, +0.79] \\
MEC$^4$ & Qwen & 20 & 8092.0 & $49.85_{\scriptscriptstyle\pm0.08}$ & $49.85_{\scriptscriptstyle\pm0.08}$ & $49.66_{\scriptscriptstyle\pm0.18}$ & $\mathbf{49.91}_{\scriptscriptstyle\pm0.14}$ & +0.25 [-0.13, +0.62] \\
MEC$^4$ & Gemma & 5 & 2023.0 & $45.84_{\scriptscriptstyle\pm0.38}$ & $45.82_{\scriptscriptstyle\pm0.47}$ & $48.09_{\scriptscriptstyle\pm0.94}$ & $\mathbf{48.38}_{\scriptscriptstyle\pm0.31}$ & +0.29 [-0.25, +0.82] \\
MEC$^4$ & Gemma & 10 & 4046.0 & $49.26_{\scriptscriptstyle\pm0.28}$ & $49.26_{\scriptscriptstyle\pm0.28}$ & $49.25_{\scriptscriptstyle\pm0.54}$ & $\mathbf{49.63}_{\scriptscriptstyle\pm0.32}$ & +0.38 [-0.14, +0.92] \\
MEC$^4$ & Gemma & 20 & 8092.0 & $50.27_{\scriptscriptstyle\pm0.06}$ & $50.27_{\scriptscriptstyle\pm0.06}$ & $49.92_{\scriptscriptstyle\pm0.26}$ & $\mathbf{50.32}_{\scriptscriptstyle\pm0.11}$ & +0.40 [-0.08, +0.85] \\
\bottomrule
\end{tabular}
\end{table*}

\section{Query Sides and Composition Controls}
\label{app:merged-F}
\label{sec:s-side}
\subsection{Why retain both sides of the boundary?}
\label{app:merged-F-1}
Negative-only (rejected) and positive-only (accepted) policies rank eligible relations by distance to the student threshold. The restricted policy for rejected candidates retains at most the nearest one for each target. If a dialogue has fewer eligible relations than its allowance, both the restricted selector and its bilateral comparator use that smaller count.

At the primary 10\% allowance, every paired interval favors querying both sides of the threshold (Table~\ref{tab:s-side-10}). Restricting queries to rejected predictions leaves false positives untouched. Restricting them to accepted predictions leaves missed causes unrecovered. Limiting queries to one rejected candidate per target also loses in all four settings at the same actual answer count.

\begin{table*}[!tbp]
\centering
\caption{Query side at nominal 10\% access under fixed fusion. Both methods use the actual $Q$ shown. $\Delta$ is Bilateral minus the restricted selector.}
\label{tab:s-side-10}
\begin{tabular}{@{}lllrrrr@{}}
\toprule
Dataset & Model & Selector & $Q$ & Restricted & Bilateral & $\Delta$ [95\% CI] \\
\midrule
ECF & Qwen & Negative only & 1490.3 & $63.68_{\scriptscriptstyle\pm0.35}$ & $64.89_{\scriptscriptstyle\pm0.79}$ & +1.21 [+0.76, +1.61] \\
ECF & Qwen & Positive only & 1402.3 & $63.79_{\scriptscriptstyle\pm1.38}$ & $64.88_{\scriptscriptstyle\pm0.76}$ & +1.09 [+0.54, +1.62] \\
ECF & Qwen & One rejected/target & 1464.3 & $63.61_{\scriptscriptstyle\pm0.37}$ & $64.85_{\scriptscriptstyle\pm0.77}$ & +1.24 [+0.79, +1.66] \\
ECF & Gemma & Negative only & 1453.3 & $64.03_{\scriptscriptstyle\pm0.37}$ & $65.14_{\scriptscriptstyle\pm0.28}$ & +1.10 [+0.50, +1.72] \\
ECF & Gemma & Positive only & 1351.7 & $62.96_{\scriptscriptstyle\pm1.09}$ & $64.99_{\scriptscriptstyle\pm0.32}$ & +2.03 [+1.36, +2.66] \\
ECF & Gemma & One rejected/target & 1427.7 & $63.41_{\scriptscriptstyle\pm0.21}$ & $65.10_{\scriptscriptstyle\pm0.32}$ & +1.69 [+1.07, +2.30] \\
MEC$^4$ & Qwen & Negative only & 3906.3 & $48.50_{\scriptscriptstyle\pm0.95}$ & $49.74_{\scriptscriptstyle\pm0.60}$ & +1.24 [+0.67, +1.77] \\
MEC$^4$ & Qwen & Positive only & 2663.0 & $48.45_{\scriptscriptstyle\pm0.59}$ & $49.29_{\scriptscriptstyle\pm0.65}$ & +0.84 [+0.33, +1.38] \\
MEC$^4$ & Qwen & One rejected/target & 3809.0 & $48.36_{\scriptscriptstyle\pm0.96}$ & $49.69_{\scriptscriptstyle\pm0.59}$ & +1.33 [+0.76, +1.89] \\
MEC$^4$ & Gemma & Negative only & 3905.0 & $47.52_{\scriptscriptstyle\pm0.67}$ & $49.02_{\scriptscriptstyle\pm0.86}$ & +1.51 [+0.97, +2.03] \\
MEC$^4$ & Gemma & Positive only & 2835.3 & $47.33_{\scriptscriptstyle\pm0.15}$ & $48.41_{\scriptscriptstyle\pm0.64}$ & +1.08 [+0.54, +1.64] \\
MEC$^4$ & Gemma & One rejected/target & 3829.3 & $47.35_{\scriptscriptstyle\pm0.57}$ & $49.00_{\scriptscriptstyle\pm0.83}$ & +1.65 [+1.11, +2.22] \\
\bottomrule
\end{tabular}
\end{table*}

\subsection{Smaller and larger allowances}
\label{app:merged-F-2}
Tables~\ref{tab:s-side-5} and~\ref{tab:s-side-20} examine this pattern at other budgets. At 5\%, bilateral selection has a higher mean in every comparison. Its smallest difference is against positive-only Qwen/ECF, at 0.28 F1 with interval $[-0.15,0.74]$. At 20\%, every interval favors bilateral selection. Access to both repair directions becomes more valuable as the allowance grows.

At the wider allowance, the restricted policies often run out of eligible candidates, particularly on MEC$^4$. The bilateral comparator is reduced to the same actual count. Its remaining advantage reflects which errors can be addressed with those answers.

\begin{table*}[!tbp]
\centering
\caption{Query side at nominal 5\% access under fixed fusion. Both methods use the actual $Q$ shown. $\Delta$ is Bilateral minus the restricted selector.}
\label{tab:s-side-5}
\begin{tabular}{@{}lllrrrr@{}}
\toprule
Dataset & Model & Selector & $Q$ & Restricted & Bilateral & $\Delta$ [95\% CI] \\
\midrule
ECF & Qwen & Negative only & 661.3 & $63.21_{\scriptscriptstyle\pm0.43}$ & $63.69_{\scriptscriptstyle\pm0.63}$ & +0.48 [+0.13, +0.82] \\
ECF & Qwen & Positive only & 653.3 & $63.41_{\scriptscriptstyle\pm1.22}$ & $63.69_{\scriptscriptstyle\pm0.63}$ & +0.28 [-0.15, +0.74] \\
ECF & Qwen & One rejected/target & 660.0 & $63.27_{\scriptscriptstyle\pm0.44}$ & $63.69_{\scriptscriptstyle\pm0.62}$ & +0.42 [+0.07, +0.75] \\
ECF & Gemma & Negative only & 713.3 & $63.23_{\scriptscriptstyle\pm0.19}$ & $63.95_{\scriptscriptstyle\pm0.20}$ & +0.71 [+0.26, +1.17] \\
ECF & Gemma & Positive only & 701.3 & $62.70_{\scriptscriptstyle\pm1.05}$ & $63.96_{\scriptscriptstyle\pm0.21}$ & +1.26 [+0.78, +1.72] \\
ECF & Gemma & One rejected/target & 712.3 & $63.05_{\scriptscriptstyle\pm0.21}$ & $63.95_{\scriptscriptstyle\pm0.20}$ & +0.90 [+0.43, +1.35] \\
MEC$^4$ & Qwen & Negative only & 2004.7 & $48.35_{\scriptscriptstyle\pm1.00}$ & $49.10_{\scriptscriptstyle\pm0.74}$ & +0.76 [+0.29, +1.20] \\
MEC$^4$ & Qwen & Positive only & 1856.0 & $48.40_{\scriptscriptstyle\pm0.61}$ & $48.95_{\scriptscriptstyle\pm0.71}$ & +0.55 [+0.07, +1.08] \\
MEC$^4$ & Qwen & One rejected/target & 2004.7 & $48.32_{\scriptscriptstyle\pm0.92}$ & $49.10_{\scriptscriptstyle\pm0.74}$ & +0.79 [+0.31, +1.24] \\
MEC$^4$ & Gemma & Negative only & 1854.7 & $47.04_{\scriptscriptstyle\pm0.61}$ & $48.08_{\scriptscriptstyle\pm0.20}$ & +1.04 [+0.55, +1.48] \\
MEC$^4$ & Gemma & Positive only & 1777.3 & $47.11_{\scriptscriptstyle\pm0.27}$ & $47.98_{\scriptscriptstyle\pm0.13}$ & +0.88 [+0.42, +1.33] \\
MEC$^4$ & Gemma & One rejected/target & 1854.7 & $46.99_{\scriptscriptstyle\pm0.54}$ & $48.08_{\scriptscriptstyle\pm0.20}$ & +1.10 [+0.60, +1.54] \\
\bottomrule
\end{tabular}
\end{table*}
\begin{table*}[!tbp]
\centering
\caption{Query side at nominal 20\% access under fixed fusion. Both methods use the actual $Q$ shown. $\Delta$ is Bilateral minus the restricted selector.}
\label{tab:s-side-20}
\begin{tabular}{@{}lllrrrr@{}}
\toprule
Dataset & Model & Selector & $Q$ & Restricted & Bilateral & $\Delta$ [95\% CI] \\
\midrule
ECF & Qwen & Negative only & 3107.7 & $63.76_{\scriptscriptstyle\pm0.24}$ & $65.51_{\scriptscriptstyle\pm0.80}$ & +1.75 [+1.29, +2.15] \\
ECF & Qwen & Positive only & 2088.0 & $63.82_{\scriptscriptstyle\pm1.39}$ & $65.16_{\scriptscriptstyle\pm0.90}$ & +1.34 [+0.73, +1.91] \\
ECF & Qwen & One rejected/target & 2373.7 & $63.61_{\scriptscriptstyle\pm0.32}$ & $65.24_{\scriptscriptstyle\pm0.72}$ & +1.63 [+1.20, +2.06] \\
ECF & Gemma & Negative only & 3092.0 & $64.20_{\scriptscriptstyle\pm0.26}$ & $65.98_{\scriptscriptstyle\pm0.35}$ & +1.78 [+1.19, +2.38] \\
ECF & Gemma & Positive only & 1912.0 & $62.99_{\scriptscriptstyle\pm1.01}$ & $65.35_{\scriptscriptstyle\pm0.43}$ & +2.36 [+1.68, +3.04] \\
ECF & Gemma & One rejected/target & 2365.3 & $63.45_{\scriptscriptstyle\pm0.11}$ & $65.83_{\scriptscriptstyle\pm0.45}$ & +2.38 [+1.75, +3.02] \\
MEC$^4$ & Qwen & Negative only & 8314.0 & $48.52_{\scriptscriptstyle\pm0.93}$ & $49.90_{\scriptscriptstyle\pm0.54}$ & +1.38 [+0.80, +1.91] \\
MEC$^4$ & Qwen & Positive only & 2809.3 & $48.46_{\scriptscriptstyle\pm0.58}$ & $49.30_{\scriptscriptstyle\pm0.64}$ & +0.83 [+0.31, +1.36] \\
MEC$^4$ & Qwen & One rejected/target & 5642.7 & $48.36_{\scriptscriptstyle\pm0.93}$ & $49.76_{\scriptscriptstyle\pm0.57}$ & +1.40 [+0.80, +1.95] \\
MEC$^4$ & Gemma & Negative only & 8140.3 & $47.54_{\scriptscriptstyle\pm0.67}$ & $49.45_{\scriptscriptstyle\pm1.01}$ & +1.91 [+1.38, +2.45] \\
MEC$^4$ & Gemma & Positive only & 3008.3 & $47.35_{\scriptscriptstyle\pm0.11}$ & $48.43_{\scriptscriptstyle\pm0.59}$ & +1.08 [+0.53, +1.66] \\
MEC$^4$ & Gemma & One rejected/target & 5733.3 & $47.34_{\scriptscriptstyle\pm0.60}$ & $49.25_{\scriptscriptstyle\pm0.94}$ & +1.91 [+1.36, +2.47] \\
\bottomrule
\end{tabular}
\end{table*}

\subsection{Total access versus dialogue allocation}
\label{app:merged-F-3}
Table~\ref{tab:s-random} varies how random queries are matched. Uniform matched total preserves the selective policy's total query count, uniform nominal uses the nominal fraction over all candidates, and per-dialogue matched preserves each dialogue's actual count. All three use the same 20 random seeds, frozen calibration, and response rules.

The selective advantage persists across the random controls in Table~\ref{tab:s-random}. Sharing the selective policy's dialogue quotas can help random consultation because those counts indicate where the student seeks assistance. Even with that allocation information available, random queries make less effective choices within each dialogue.

\begin{table*}[!tbp]
\centering
\caption{Additional random query controls on test. Query fractions and F1 are group means, with $\pm$ denoting sample SD across three students. All policies use the same 20 random query seeds.}
\label{tab:s-random}
\begin{tabular}{llrlrrr}
\toprule
Dataset & Model & \shortstack{Nominal\\query \%} & Random policy & \shortstack{Actual\\query \%} & Fusion & Replacement \\
\midrule
ECF & Qwen & 5 & uniform matched total & 4.13 & $62.15_{\scriptscriptstyle\pm0.77}$ & $62.20_{\scriptscriptstyle\pm0.72}$ \\
ECF & Qwen & 5 & uniform nominal & 5.00 & $62.17_{\scriptscriptstyle\pm0.76}$ & $62.23_{\scriptscriptstyle\pm0.71}$ \\
ECF & Qwen & 5 & per-dialogue matched & 4.13 & $62.23_{\scriptscriptstyle\pm0.74}$ & $62.29_{\scriptscriptstyle\pm0.69}$ \\
ECF & Qwen & 10 & uniform matched total & 9.31 & $62.31_{\scriptscriptstyle\pm0.76}$ & $62.43_{\scriptscriptstyle\pm0.67}$ \\
ECF & Qwen & 10 & uniform nominal & 10.00 & $62.33_{\scriptscriptstyle\pm0.77}$ & $62.45_{\scriptscriptstyle\pm0.66}$ \\
ECF & Qwen & 10 & per-dialogue matched & 9.31 & $62.41_{\scriptscriptstyle\pm0.79}$ & $62.52_{\scriptscriptstyle\pm0.66}$ \\
ECF & Qwen & 20 & uniform matched total & 19.41 & $62.67_{\scriptscriptstyle\pm0.76}$ & $62.88_{\scriptscriptstyle\pm0.57}$ \\
ECF & Qwen & 20 & uniform nominal & 20.00 & $62.69_{\scriptscriptstyle\pm0.78}$ & $62.91_{\scriptscriptstyle\pm0.57}$ \\
ECF & Qwen & 20 & per-dialogue matched & 19.41 & $62.81_{\scriptscriptstyle\pm0.83}$ & $63.03_{\scriptscriptstyle\pm0.66}$ \\
ECF & Gemma & 5 & uniform matched total & 4.46 & $61.49_{\scriptscriptstyle\pm0.62}$ & $61.56_{\scriptscriptstyle\pm0.61}$ \\
ECF & Gemma & 5 & uniform nominal & 5.00 & $61.51_{\scriptscriptstyle\pm0.62}$ & $61.58_{\scriptscriptstyle\pm0.61}$ \\
ECF & Gemma & 5 & per-dialogue matched & 4.46 & $61.50_{\scriptscriptstyle\pm0.62}$ & $61.52_{\scriptscriptstyle\pm0.59}$ \\
ECF & Gemma & 10 & uniform matched total & 9.08 & $61.72_{\scriptscriptstyle\pm0.57}$ & $61.81_{\scriptscriptstyle\pm0.55}$ \\
ECF & Gemma & 10 & uniform nominal & 10.00 & $61.76_{\scriptscriptstyle\pm0.58}$ & $61.84_{\scriptscriptstyle\pm0.57}$ \\
ECF & Gemma & 10 & per-dialogue matched & 9.08 & $61.79_{\scriptscriptstyle\pm0.59}$ & $61.86_{\scriptscriptstyle\pm0.56}$ \\
ECF & Gemma & 20 & uniform matched total & 19.32 & $62.18_{\scriptscriptstyle\pm0.51}$ & $62.34_{\scriptscriptstyle\pm0.48}$ \\
ECF & Gemma & 20 & uniform nominal & 20.00 & $62.21_{\scriptscriptstyle\pm0.52}$ & $62.38_{\scriptscriptstyle\pm0.48}$ \\
ECF & Gemma & 20 & per-dialogue matched & 19.32 & $62.24_{\scriptscriptstyle\pm0.56}$ & $62.36_{\scriptscriptstyle\pm0.42}$ \\
MEC$^4$ & Qwen & 5 & uniform matched total & 4.95 & $47.23_{\scriptscriptstyle\pm0.61}$ & $47.21_{\scriptscriptstyle\pm0.59}$ \\
MEC$^4$ & Qwen & 5 & uniform nominal & 5.00 & $47.23_{\scriptscriptstyle\pm0.61}$ & $47.21_{\scriptscriptstyle\pm0.59}$ \\
MEC$^4$ & Qwen & 5 & per-dialogue matched & 4.95 & $47.29_{\scriptscriptstyle\pm0.61}$ & $47.30_{\scriptscriptstyle\pm0.61}$ \\
MEC$^4$ & Qwen & 10 & uniform matched total & 9.65 & $47.35_{\scriptscriptstyle\pm0.61}$ & $47.28_{\scriptscriptstyle\pm0.58}$ \\
MEC$^4$ & Qwen & 10 & uniform nominal & 10.00 & $47.37_{\scriptscriptstyle\pm0.60}$ & $47.30_{\scriptscriptstyle\pm0.57}$ \\
MEC$^4$ & Qwen & 10 & per-dialogue matched & 9.65 & $47.44_{\scriptscriptstyle\pm0.63}$ & $47.46_{\scriptscriptstyle\pm0.61}$ \\
MEC$^4$ & Qwen & 20 & uniform matched total & 20.55 & $47.65_{\scriptscriptstyle\pm0.58}$ & $47.62_{\scriptscriptstyle\pm0.48}$ \\
MEC$^4$ & Qwen & 20 & uniform nominal & 20.00 & $47.64_{\scriptscriptstyle\pm0.59}$ & $47.60_{\scriptscriptstyle\pm0.49}$ \\
MEC$^4$ & Qwen & 20 & per-dialogue matched & 20.55 & $47.74_{\scriptscriptstyle\pm0.55}$ & $47.78_{\scriptscriptstyle\pm0.45}$ \\
MEC$^4$ & Gemma & 5 & uniform matched total & 4.58 & $45.67_{\scriptscriptstyle\pm0.27}$ & $45.68_{\scriptscriptstyle\pm0.29}$ \\
MEC$^4$ & Gemma & 5 & uniform nominal & 5.00 & $45.69_{\scriptscriptstyle\pm0.27}$ & $45.70_{\scriptscriptstyle\pm0.29}$ \\
MEC$^4$ & Gemma & 5 & per-dialogue matched & 4.58 & $45.68_{\scriptscriptstyle\pm0.27}$ & $45.69_{\scriptscriptstyle\pm0.28}$ \\
MEC$^4$ & Gemma & 10 & uniform matched total & 9.65 & $45.87_{\scriptscriptstyle\pm0.23}$ & $45.91_{\scriptscriptstyle\pm0.30}$ \\
MEC$^4$ & Gemma & 10 & uniform nominal & 10.00 & $45.88_{\scriptscriptstyle\pm0.21}$ & $45.92_{\scriptscriptstyle\pm0.29}$ \\
MEC$^4$ & Gemma & 10 & per-dialogue matched & 9.65 & $45.92_{\scriptscriptstyle\pm0.16}$ & $45.95_{\scriptscriptstyle\pm0.25}$ \\
MEC$^4$ & Gemma & 20 & uniform matched total & 20.12 & $46.31_{\scriptscriptstyle\pm0.15}$ & $46.44_{\scriptscriptstyle\pm0.26}$ \\
MEC$^4$ & Gemma & 20 & uniform nominal & 20.00 & $46.31_{\scriptscriptstyle\pm0.13}$ & $46.43_{\scriptscriptstyle\pm0.22}$ \\
MEC$^4$ & Gemma & 20 & per-dialogue matched & 20.12 & $46.24_{\scriptscriptstyle\pm0.21}$ & $46.27_{\scriptscriptstyle\pm0.37}$ \\
\bottomrule
\relax

\end{tabular}
\end{table*}

\subsection{Matching predicted class and position}
\label{app:merged-F-4}
Table~\ref{tab:s-stratified} further matches queries jointly by dialogue, the class predicted by the student, and four distance categories (self, previous turn, earlier, and future). Twenty random draws preserve each stratum's count. This holds the proportions of relation types fixed while changing selection within each type.

Selective consultation exceeds stratified random selection by 1.30--2.37 F1 points across students, with a positive paired interval for every student. Matching class and position composition therefore leaves an advantage for boundary ranking within those groups.

\begin{table*}[!tbp]
\centering
\caption{Stratified random control at nominal 10\% access. Twenty random draws preserve counts jointly by dialogue, the student's predicted class, and distance category (self, previous, earlier, or future). $\Delta$ denotes Boundary minus Stratified.}
\label{tab:s-stratified}
\begin{tabular}{@{}llrrrrr@{}}
\toprule
Dataset & Model & Seed & $Q$ & Boundary & Stratified & $\Delta$ [95\% CI] \\
\midrule
ECF & Qwen & 42 & 1544 & 64.00 & 62.35 & +1.65 [+0.88, +2.37] \\
ECF & Qwen & 43 & 1447 & 65.11 & 63.12 & +1.98 [+1.39, +2.57] \\
ECF & Qwen & 44 & 1480 & 65.55 & 64.10 & +1.44 [+0.87, +2.02] \\
ECF & Gemma & 42 & 1428 & 64.88 & 62.93 & +1.95 [+1.36, +2.56] \\
ECF & Gemma & 43 & 1395 & 65.46 & 63.72 & +1.74 [+1.09, +2.37] \\
ECF & Gemma & 44 & 1539 & 65.09 & 62.72 & +2.37 [+1.49, +3.22] \\
MEC$^4$ & Qwen & 42 & 4104 & 50.27 & 48.97 & +1.30 [+0.73, +1.88] \\
MEC$^4$ & Qwen & 43 & 3877 & 49.87 & 48.26 & +1.61 [+0.98, +2.21] \\
MEC$^4$ & Qwen & 44 & 3738 & 49.09 & 47.76 & +1.32 [+0.71, +1.94] \\
MEC$^4$ & Gemma & 42 & 4087 & 49.18 & 47.28 & +1.90 [+1.30, +2.50] \\
MEC$^4$ & Gemma & 43 & 3802 & 49.80 & 47.57 & +2.22 [+1.51, +2.89] \\
MEC$^4$ & Gemma & 44 & 3826 & 48.10 & 46.79 & +1.30 [+0.72, +1.87] \\
\bottomrule
\end{tabular}
\end{table*}

\section{Response Integration}
\label{app:merged-G}
\label{sec:s-integration}
\subsection{Response rules and access endpoints}
\label{app:merged-G-1}
Table~\ref{tab:s-integration-summary} compares consultation at nominal 20\% access with the independent expert. Replacement adopts the expert's calibrated decision, whereas fusion can retain a confident student judgment even when the expert disagrees.

\begin{table}[!htbp]
\centering
\caption{Integration at nominal 20\% access, mean$_{\scriptscriptstyle\pm\mathrm{SD}}$ test F1. F denotes fixed fusion, Local F fits weights and thresholds for queried relations, and R denotes replacement. Expert uses all relations. Row maxima are bold.}
\label{tab:s-integration-summary}
\begin{tabular}{@{}llrrrr@{}}
\toprule
Dataset & Model & F & Local F & R & Expert \\
\midrule
ECF & Qwen & $65.51_{\scriptscriptstyle\pm0.80}$ & $66.31_{\scriptscriptstyle\pm0.25}$ & $66.33_{\scriptscriptstyle\pm0.05}$ & $\mathbf{66.60}$ \\
ECF & Gemma & $65.99_{\scriptscriptstyle\pm0.36}$ & $66.58_{\scriptscriptstyle\pm0.28}$ & $66.43_{\scriptscriptstyle\pm0.19}$ & $\mathbf{66.70}$ \\
MEC$^4$ & Qwen & $49.90_{\scriptscriptstyle\pm0.54}$ & $\mathbf{50.28}_{\scriptscriptstyle\pm0.08}$ & $49.86_{\scriptscriptstyle\pm0.09}$ & $49.79$ \\
MEC$^4$ & Gemma & $49.45_{\scriptscriptstyle\pm1.01}$ & $\mathbf{50.28}_{\scriptscriptstyle\pm0.25}$ & $50.23_{\scriptscriptstyle\pm0.04}$ & $50.27$ \\
\bottomrule
\end{tabular}
\end{table}

\subsection{Fitting the queried region}
\label{app:merged-G-2}
Local fusion uses $(1-\alpha)z_s+\alpha z_e>\tau_q$ only on queried relations. Validation searches $\alpha\in\{0.25,0.5,0.75\}$ and the common threshold grid, also including the student threshold. Equal F1 favors a weight nearer $0.5$, then a threshold nearer $\tau_s$. Table~\ref{tab:s-local-config} reports the selected parameters and test F1 at 20\%. The configuration file retains full precision.

Most students select a larger expert weight, and Local F improves or preserves every result from fixed fusion. Equal logit weights can therefore underuse information already present in the selected answers.

\begin{table}[!htbp]
\centering
\caption{Validation-selected Local F parameters and test F1 at nominal 20\% access. F uses $\alpha=0.5$ and $\tau_s$. Local F uses the listed parameters.}
\label{tab:s-local-config}
\begin{tabular}{@{}llrrrrr@{}}
\toprule
Dataset & Model & Seed & $\alpha$ & $\tau_q$ & F & Local F \\
\midrule
ECF & Qwen & 42 & 0.75 & -0.65 & 64.59 & 66.02 \\
ECF & Qwen & 43 & 0.75 & -0.75 & 65.87 & 66.50 \\
ECF & Qwen & 44 & 0.50 & -0.85 & 66.05 & 66.40 \\
ECF & Gemma & 42 & 0.75 & -0.65 & 65.69 & 66.28 \\
ECF & Gemma & 43 & 0.75 & -0.55 & 66.38 & 66.61 \\
ECF & Gemma & 44 & 0.75 & -0.70 & 65.89 & 66.84 \\
MEC$^4$ & Qwen & 42 & 0.50 & -1.20 & 50.32 & 50.32 \\
MEC$^4$ & Qwen & 43 & 0.75 & -1.25 & 50.10 & 50.20 \\
MEC$^4$ & Qwen & 44 & 0.75 & -1.20 & 49.29 & 50.33 \\
MEC$^4$ & Gemma & 42 & 0.75 & -1.15 & 49.48 & 50.23 \\
MEC$^4$ & Gemma & 43 & 0.75 & -1.35 & 50.44 & 50.55 \\
MEC$^4$ & Gemma & 44 & 0.75 & -1.35 & 48.42 & 50.05 \\
\bottomrule
\end{tabular}
\end{table}

\section{Online Execution and Resource Use}
\label{app:merged-H}
\label{sec:s-timing}
\subsection{Timing protocol and model residency}
\label{app:merged-H-1}
Timing uses the same 32 ECF validation dialogues for both families, selected by evenly spaced dialogue-length order statistics without using labels or prediction quality. They contain 1,847 candidate relations. Each family runs alone on one NVIDIA A100-SXM4 80 GB GPU. All seven policies first execute the entire panel as warmup. Three subsequent sweeps rotate policy order. Every measured forward shape must have appeared during warmup.

The timed interval includes prompt processing, student scoring, selection, requested expert scoring, fusion, and required weight transfers. Loading weights from storage and preparing the initial resident service occur beforehand. Qwen keeps both models resident for consultation and shares an exact dialogue prefix within each model. Its batch limits are 32 student suffixes and eight expert suffixes, also bounded by 8,192 padded suffix tokens. Prefix caches are rebuilt for every dialogue and run.

Gemma uses native independent batches of two student relations or one expert relation without prefix reuse. It alternates student and expert residency. Consultation includes transfers in both directions and ends with the student ready for the next panel. Independent expert timing begins with the expert resident. No result cache is reused across policies or repeats.

Fewer expert answers reduce execution time even though most dialogues still receive queries (Table~\ref{tab:s-timing}). Processing their prefixes and managing model residency continue to cost time, so runtime does not fall in proportion to the query fraction. Random consultation with matched access takes similar time.

\begin{table}[!htbp]
\centering
\caption{Online execution on 1,847 relations. Time is mean$_{\scriptscriptstyle\pm\mathrm{SD}}$ over three warmed repeats. $D$ counts dialogues with at least one expert query out of 32. Tokens count processed input positions, including padding and repeated prefixes. k denotes thousands.}
\label{tab:s-timing}
\begin{tabular}{@{}lrrrrr@{}}
\toprule
Policy & $q$ (\%) & $D$ & Time (s) & Tok. (k) & Pairs/s \\
\midrule
\multicolumn{6}{l}{\textit{Qwen}} \\
Student & 0.00 & 0.0 & $15.57_{\scriptscriptstyle\pm0.10}$ & 181.5 & 118.7 \\
Expert & 100.00 & 32.0 & $61.58_{\scriptscriptstyle\pm0.16}$ & 168.5 & 30.0 \\
Full fusion & 100.00 & 32.0 & $76.88_{\scriptscriptstyle\pm0.07}$ & 350.0 & 24.0 \\
Sel. 5\% & 3.95 & 25.0 & $23.71_{\scriptscriptstyle\pm0.01}$ & 195.6 & 77.9 \\
Sel. 10\% & 9.42 & 30.0 & $27.60_{\scriptscriptstyle\pm0.02}$ & 206.6 & 66.9 \\
Sel. 20\% & 18.79 & 31.0 & $32.93_{\scriptscriptstyle\pm0.02}$ & 222.5 & 56.1 \\
Rnd. 10\% & 9.42 & 30.0 & $27.40_{\scriptscriptstyle\pm0.01}$ & 205.8 & 67.4 \\
\addlinespace
\multicolumn{6}{l}{\textit{Gemma}} \\
Student & 0.00 & 0.0 & $70.49_{\scriptscriptstyle\pm0.14}$ & 724.9 & 26.2 \\
Expert & 100.00 & 32.0 & $338.33_{\scriptscriptstyle\pm0.38}$ & 732.1 & 5.5 \\
Full fusion & 100.00 & 32.0 & $461.67_{\scriptscriptstyle\pm1.66}$ & 1457.0 & 4.0 \\
Sel. 5\% & 4.60 & 27.0 & $139.10_{\scriptscriptstyle\pm2.76}$ & 758.0 & 13.3 \\
Sel. 10\% & 8.55 & 29.0 & $154.27_{\scriptscriptstyle\pm0.11}$ & 788.3 & 12.0 \\
Sel. 20\% & 18.95 & 31.0 & $189.44_{\scriptscriptstyle\pm0.26}$ & 864.3 & 9.7 \\
Rnd. 10\% & 8.55 & 29.0 & $151.39_{\scriptscriptstyle\pm2.88}$ & 788.0 & 12.2 \\
\bottomrule
\end{tabular}
\end{table}

\subsection{Memory and online prediction agreement}
\label{app:merged-H-2}
Peak memory depends more directly on model residency than on the number of queried relations (Table~\ref{tab:s-timing-memory}). Qwen keeps both models resident, while Gemma alternates them. Reducing queries does not remove the need to hold a model's weights when it executes.

The Diff column compares online predictions with replay of cached scores at identical query positions. Small execution differences occur, including eight student differences for Gemma. All unqueried decisions exactly match the corresponding online student predictions.

\begin{table}[!htbp]
\centering
\caption{Peak PyTorch allocated and reserved GPU memory in GiB over the three formal repeats. Diff counts final predictions differing from cached-score replay at the same queries. No unqueried decision changes relative to the online student.}
\label{tab:s-timing-memory}
\begin{tabular}{@{}llrrr@{}}
\toprule
Model & Policy & Allocated & Reserved & Diff \\
\midrule
Qwen & Student & 10.66 & 16.61 & 0 \\
Qwen & Expert & 52.05 & 53.36 & 0 \\
Qwen & Full fusion & 60.83 & 67.48 & 0 \\
Qwen & Selective 5\% & 60.83 & 67.48 & 2 \\
Qwen & Selective 10\% & 60.83 & 67.48 & 3 \\
Qwen & Selective 20\% & 60.83 & 67.48 & 2 \\
Qwen & Random 10\% & 60.83 & 67.48 & 0 \\
Gemma & Student & 14.14 & 15.18 & 8 \\
Gemma & Expert & 57.64 & 58.16 & 10 \\
Gemma & Full fusion & 57.64 & 58.16 & 5 \\
Gemma & Selective 5\% & 57.61 & 58.05 & 4 \\
Gemma & Selective 10\% & 57.63 & 58.14 & 4 \\
Gemma & Selective 20\% & 57.63 & 58.15 & 5 \\
Gemma & Random 10\% & 57.61 & 58.05 & 8 \\
\bottomrule
\end{tabular}
\end{table}

\section{Location and Specificity of Expert Revisions}
\label{app:merged-I}
\label{sec:s-revisions}
\subsection{Net value across boundary-distance bands}
\label{app:merged-I-1}
Table~\ref{tab:s-margin-full} reports repairs and damage per 100 queried relations in distance bins defined on validation. A repair corrects a student error. Damage turns a correct student decision into an error. Their difference gives net correct revisions per 100 queries. The table pools correction counts across three students and reports the mean bin size.

Near the boundary, answers often change predictions and repairs outnumber damage (Table~\ref{tab:s-margin-full}). Farther away, fusion usually leaves the prediction unchanged because a larger adjustment is required to cross the student's threshold.

The repair counts also distinguish precision and recall effects. At nominal 10\%, Qwen/ECF fusion adds 84.7 true positives and removes 177.7 false positives on average, while losing 37.0 true positives and introducing 99.0 false positives. Net correct revisions count these changes equally. Pair F1 weights their effect through the existing true positive, false positive, and gold counts.

Replacement changes more confident predictions farther from the boundary, including correct student decisions. For Qwen/MEC$^4$, the 10--20\% bin has 3.30 repairs and 5.16 damaging revisions per 100 queries. This negative balance explains why widening access can reduce the benefit of replacement. The F1 effect also depends on whether each revision changes a true positive or false positive.

\begin{table*}[!tbp]
\centering
\caption{Repairs and damage per 100 queries by validation distance bin. Counts are pooled across three seeds and $Q$ is mean bin size. F denotes fixed fusion and R replacement.}
\label{tab:s-margin-full}
\begin{tabular}{@{}lllrrrrr@{}}
\toprule
Dataset & Model & Bin (\%) & $Q$ & F repair & F damage & R repair & R damage \\
\midrule
ECF & Qwen & 0--5 & 661.3 & 23.59 & 13.16 & 26.61 & 15.02 \\
ECF & Qwen & 5--10 & 829.0 & 12.83 & 5.91 & 20.59 & 11.06 \\
ECF & Qwen & 10--20 & 1618.0 & 2.58 & 1.26 & 8.47 & 6.26 \\
ECF & Qwen & 20--100 & 12907.7 & 0.02 & 0.01 & 0.47 & 0.50 \\
ECF & Gemma & 0--5 & 714.0 & 27.78 & 11.53 & 30.44 & 14.19 \\
ECF & Gemma & 5--10 & 740.0 & 12.57 & 6.17 & 19.41 & 12.25 \\
ECF & Gemma & 10--20 & 1640.0 & 3.37 & 1.50 & 9.07 & 6.48 \\
ECF & Gemma & 20--100 & 12922.0 & 0.02 & 0.00 & 0.42 & 0.50 \\
MEC$^4$ & Qwen & 0--5 & 2004.7 & 18.21 & 13.19 & 22.07 & 17.69 \\
MEC$^4$ & Qwen & 5--10 & 1901.7 & 6.29 & 3.33 & 13.76 & 10.94 \\
MEC$^4$ & Qwen & 10--20 & 4408.0 & 0.46 & 0.36 & 3.30 & 5.16 \\
MEC$^4$ & Qwen & 20--100 & 32147.7 & 0.00 & 0.00 & 0.06 & 0.14 \\
MEC$^4$ & Gemma & 0--5 & 1854.7 & 22.09 & 13.32 & 29.80 & 13.17 \\
MEC$^4$ & Gemma & 5--10 & 2050.3 & 6.18 & 4.29 & 15.77 & 8.23 \\
MEC$^4$ & Gemma & 10--20 & 4236.3 & 1.01 & 0.54 & 6.33 & 3.50 \\
MEC$^4$ & Gemma & 20--100 & 32320.7 & 0.00 & 0.00 & 0.06 & 0.05 \\
\bottomrule
\end{tabular}
\end{table*}

\subsection{Full fusion outside the query band}
\label{app:merged-I-2}
For all six MEC$^4$ students, 20\% selective fusion and full fusion at the student threshold produce identical prediction sets (Table~\ref{tab:s-outside}). The plateau reflects unchanged predictions outside the band, not offsetting repairs and damage.

On ECF, the few differences outside the band include both repairs and damage. Full access recovers some true positives and loses others, while also removing and adding false positives. The small net F1 change combines those opposing revisions. Replacement can affect more decisions because it does not average the expert score with the student score.

\begin{table*}[!tbp]
\centering
\begin{minipage}[t]{0.48\textwidth}
\centering
\caption{Prediction differences between 20\% selective fusion and full fusion, both at $\tau_s$, and flips outside the query band under full fusion. Diff. counts differing predictions per student. $+\mathrm{TP}$ and $-\mathrm{FP}$ are repairs.}
\label{tab:s-outside}
\begin{tabular}{@{}llrrrrrr@{}}
\toprule
Dataset & Model & Seed & Diff. & $+\mathrm{TP}$ & $-\mathrm{TP}$ & $+\mathrm{FP}$ & $-\mathrm{FP}$ \\
\midrule
ECF & Qwen & 42 & 0 & 0 & 0 & 0 & 0 \\
ECF & Qwen & 43 & 6 & 2 & 1 & 1 & 2 \\
ECF & Qwen & 44 & 7 & 0 & 0 & 2 & 5 \\
ECF & Gemma & 42 & 4 & 2 & 0 & 0 & 2 \\
ECF & Gemma & 43 & 2 & 1 & 0 & 0 & 1 \\
ECF & Gemma & 44 & 4 & 2 & 1 & 0 & 1 \\
MEC$^4$ & Qwen & 42 & 0 & 0 & 0 & 0 & 0 \\
MEC$^4$ & Qwen & 43 & 0 & 0 & 0 & 0 & 0 \\
MEC$^4$ & Qwen & 44 & 0 & 0 & 0 & 0 & 0 \\
MEC$^4$ & Gemma & 42 & 0 & 0 & 0 & 0 & 0 \\
MEC$^4$ & Gemma & 43 & 0 & 0 & 0 & 0 & 0 \\
MEC$^4$ & Gemma & 44 & 0 & 0 & 0 & 0 & 0 \\
\bottomrule
\end{tabular}
\end{minipage}\hfill
\begin{minipage}[t]{0.48\textwidth}
\centering
\caption{Association between relations and expert responses at nominal 10\% access. Test F1 is mean$_{\scriptscriptstyle\pm\mathrm{SD}}$ across three SFT seeds. Permuted averages 50 response permutations within dialogues per student. $\Delta$ is Permuted minus Selective F.}
\label{tab:s-permutation}
\begin{tabular}{@{}llrrrr@{}}
\toprule
Dataset & Model & Student & Selective F & Permuted & $\Delta$ \\
\midrule
ECF & Qwen & $62.05_{\scriptscriptstyle\pm0.77}$ & $64.89_{\scriptscriptstyle\pm0.79}$ & $61.26_{\scriptscriptstyle\pm1.29}$ & -3.63 \\
ECF & Gemma & $61.26_{\scriptscriptstyle\pm0.66}$ & $65.14_{\scriptscriptstyle\pm0.29}$ & $60.57_{\scriptscriptstyle\pm0.93}$ & -4.57 \\
MEC$^4$ & Qwen & $47.10_{\scriptscriptstyle\pm0.64}$ & $49.74_{\scriptscriptstyle\pm0.60}$ & $44.19_{\scriptscriptstyle\pm0.90}$ & -5.55 \\
MEC$^4$ & Gemma & $45.50_{\scriptscriptstyle\pm0.31}$ & $49.02_{\scriptscriptstyle\pm0.86}$ & $43.47_{\scriptscriptstyle\pm1.03}$ & -5.55 \\
\bottomrule
\end{tabular}
\end{minipage}
\end{table*}

\subsection{Linking queries to expert answers}
\label{app:merged-I-3}
The control in Table~\ref{tab:s-permutation} keeps each dialogue's selected positions fixed and permutes the returned expert scores among them. The score multiset and query count remain unchanged. Results are first averaged over 50 permutations per student, then summarized as the mean and sample SD across the three student seeds in each setting.

Permuting responses lowers F1 below the student in every group. Useful query positions cannot compensate for answers assigned to the wrong relations. Matched random selection and response permutation test complementary requirements. The former varies where help is requested, while the latter varies whether the response addresses that request.

\section{Consultation After Further Training}
\label{app:merged-J}
\label{sec:s-kd}
\subsection{Continuing ordinary SFT}
\label{app:merged-J-1}
The ordinary continuation control completes the second pass of the original SFT schedule at seed 42, retaining its CE objective, LoRA configuration, optimizer schedule, and candidate data. The expert stays at its first-pass checkpoint from seed 42 for the same family and dataset. Calibration and query bands are refitted on validation for the continued student and frozen for evaluation. Table~\ref{tab:s-sft2} covers both model families on both datasets.

Selective fusion exceeds matched random fusion at every allowance, with positive paired intervals in all twelve comparisons. However, wider access is not uniformly better. Qwen/MEC$^4$ fusion is slightly lower at 20\% than at 10\%, whereas Gemma/MEC$^4$ and both ECF controls continue to improve.

\begin{table*}[!tbp]
\centering
\caption{Consultation after two ordinary SFT passes, seed 42, with the fixed expert from the first pass. $b$ is nominal query percentage and $Q$ actual answers. Scores are test pair F1. $\Delta$ compares selective fusion with the mean of 20 random draws matched by dialogue counts. Intervals use 2,000 paired dialogue resamples.}
\label{tab:s-sft2}
\begin{tabular}{@{}llrrrrrrr@{}}
\toprule
Dataset & Model & $b$ & $Q$ & Student & Random F & Selective F & Selective R & $\Delta$ [95\% CI] \\
\midrule
ECF & Qwen & 5 & 723 & 62.66 & 62.89 & 64.78 & 64.69 & +1.89 [+1.21, +2.58] \\
ECF & Qwen & 10 & 1,458 & 62.66 & 63.06 & 65.89 & 65.77 & +2.83 [+2.00, +3.66] \\
ECF & Qwen & 20 & 3,211 & 62.66 & 63.51 & 66.40 & 66.73 & +2.89 [+2.10, +3.67] \\
ECF & Gemma & 5 & 765 & 62.59 & 62.80 & 64.54 & 64.39 & +1.74 [+1.01, +2.47] \\
ECF & Gemma & 10 & 1,516 & 62.59 & 63.00 & 65.77 & 65.71 & +2.77 [+1.97, +3.63] \\
ECF & Gemma & 20 & 3,089 & 62.59 & 63.34 & 66.45 & 66.74 & +3.12 [+2.31, +3.93] \\
MEC$^4$ & Qwen & 5 & 2,085 & 47.73 & 47.85 & 49.33 & 50.14 & +1.48 [+0.68, +2.30] \\
MEC$^4$ & Qwen & 10 & 4,084 & 47.73 & 47.95 & 49.70 & 50.08 & +1.75 [+0.85, +2.63] \\
MEC$^4$ & Qwen & 20 & 8,478 & 47.73 & 48.21 & 49.66 & 49.80 & +1.44 [+0.64, +2.27] \\
MEC$^4$ & Gemma & 5 & 2,148 & 46.88 & 47.08 & 49.72 & 49.95 & +2.64 [+1.86, +3.43] \\
MEC$^4$ & Gemma & 10 & 4,171 & 46.88 & 47.30 & 50.16 & 50.36 & +2.86 [+1.99, +3.76] \\
MEC$^4$ & Gemma & 20 & 8,094 & 46.88 & 47.57 & 50.28 & 50.27 & +2.71 [+1.89, +3.52] \\
\bottomrule
\end{tabular}
\end{table*}

\subsection{Query ranking after ordinary continuation}
\label{app:merged-J-2}
\label{sec:s-sft2-routing}
The continued students at seed 42 are also compared using a global cap of $\lfloor0.1N\rfloor$ expert answers and fixed fusion. Every selector receives the same actual answer count and uses the existing validation calibration. The expert remains at its original first-pass checkpoint. The learned router is refitted for each continued student with the 16 features and fitting procedure in Section~\ref{sec:s-router}. Five dialogue folds select between count improvement and reduction in logistic loss. All four settings select count improvement.

Table~\ref{tab:s-sft2-routing} reports the resulting query counts and extraction scores. Learned validation scores use out-of-fold predictions, and all policies improve on the corresponding validation student. Parameters and the selected learning objective are frozen before test evaluation. The hard cap fixes the number of answers, whereas Table~\ref{tab:s-sft2} uses a frozen distance band with a nominal query rate.

\begin{table*}[!tbp]
\centering
\caption{Query rankings after two ordinary SFT passes, seed 42, under a hard global 10\% cap with fixed fusion. $Q$ is the actual number of expert answers per policy. $H_\ell$ and $H_p$ denote native and calibrated entropy. Validation uses out-of-fold predictions for Learned. Scores are pair F1, with row maxima in bold.}
\label{tab:s-sft2-routing}
\begin{tabular}{@{}llrrrrrr@{}}
\toprule
\multicolumn{8}{l}{(a) Test} \\
Dataset & Model & $Q$ & Student & $H_\ell$ & $H_p$ & Learned & Boundary \\
\midrule
ECF & Qwen & 1,601 & 62.66 & 64.40 & 64.15 & 65.16 & $\mathbf{65.98}$ \\
ECF & Gemma & 1,601 & 62.59 & 65.00 & 65.02 & 64.81 & $\mathbf{65.74}$ \\
MEC$^4$ & Qwen & 4,046 & 47.73 & 49.49 & 49.39 & 49.15 & $\mathbf{49.69}$ \\
MEC$^4$ & Gemma & 4,046 & 46.88 & 50.08 & 50.06 & 49.30 & $\mathbf{50.15}$ \\
\addlinespace
\multicolumn{8}{l}{(b) Validation} \\
Dataset & Model & $Q$ & Student & $H_\ell$ & $H_p$ & Learned & Boundary \\
ECF & Qwen & 676 & 64.04 & 65.74 & 65.66 & 66.24 & $\mathbf{67.23}$ \\
ECF & Gemma & 676 & 62.42 & 64.60 & 64.57 & 64.37 & $\mathbf{65.04}$ \\
MEC$^4$ & Qwen & 3,708 & 45.47 & 47.95 & 47.80 & 47.92 & $\mathbf{48.42}$ \\
MEC$^4$ & Gemma & 3,708 & 43.45 & 46.09 & 46.06 & 45.73 & $\mathbf{46.62}$ \\
\bottomrule
\end{tabular}
\end{table*}

Boundary retains an advantage over the refitted learned selector in all four settings (Table~\ref{tab:s-sft2-routing-intervals}). The comparison with entropy varies by dataset. On ECF, centering selection at the task threshold improves allocation over both entropy rankings. On MEC$^4$, their F1 scores are close and the paired intervals include zero.

\begin{table*}[!tbp]
\centering
\caption{Boundary minus each competing selector after two ordinary SFT passes, seed 42. All use fixed fusion and the same hard global 10\% budget. Entries are test F1 differences with pointwise 95\% paired intervals from 2,000 dialogue resamples, conditional on the trained models and frozen calibration.}
\label{tab:s-sft2-routing-intervals}
\begin{tabular}{@{}llrrr@{}}
\toprule
Dataset & Model & Boundary $-$ $H_\ell$ & Boundary $-$ $H_p$ & Boundary $-$ Learned \\
\midrule
ECF & Qwen & +1.58 [+0.84, +2.35] & +1.83 [+1.07, +2.61] & +0.82 [+0.31, +1.35] \\
ECF & Gemma & +0.74 [+0.21, +1.29] & +0.72 [+0.18, +1.25] & +0.93 [+0.35, +1.52] \\
MEC$^4$ & Qwen & +0.20 [-0.20, +0.61] & +0.30 [-0.14, +0.75] & +0.54 [+0.04, +1.06] \\
MEC$^4$ & Gemma & +0.07 [-0.29, +0.46] & +0.09 [-0.31, +0.49] & +0.85 [+0.40, +1.32] \\
\bottomrule
\end{tabular}
\end{table*}

\subsection{Expert-guided training objective}
\label{app:merged-J-3}
Each KD student starts from its own matching SFT checkpoint and receives one further pass through the same candidate data. For raw yes/no margins $\ell_s,\ell_e$, let $q=\sigma(\ell_e/T)$ and $p=\sigma(\ell_s/T)$. For a batch of $B$ relations, the auxiliary loss is
\begin{equation}
\mathcal L_{\mathrm{KD}}=\frac{T^2}{B}\sum_{r=1}^B\left[q_r\log\frac{q_r}{p_r}+(1-q_r)\log\frac{1-q_r}{1-p_r}\right].
\end{equation}
The loss is Bernoulli Kullback--Leibler (KL) divergence from teacher to student, averaged across relations, with $T=2$. The combined objective is $0.5\mathcal L_{\mathrm{CE}}+0.5\mathcal L_{\mathrm{KD}}$, where CE averages over unmasked assistant completion tokens. KD uses learning rate $2\times10^{-4}$.

\subsection{Remaining value of selective access}
\label{app:merged-J-4}
After KD, selective fusion exceeds both the student and the mean of matched random draws for every seed (Table~\ref{tab:s-kd-seeds}). Prior training with the expert therefore leaves some useful assistance at inference time. Its value depends on the remaining errors at queried relations, beyond the gap in standalone F1.

The three Qwen/MEC$^4$ intervals overlap zero, while the other nine are positive. Evidence for an allocation advantage is therefore weaker in that setting.

\begin{table*}[!tbp]
\centering
\caption{Additional KD condition at nominal 10\% access. All students have one SFT pass plus one KD pass. $\Delta$: selective fusion minus matched random fusion.}
\label{tab:s-kd-seeds}
\begin{tabular}{@{}llrrrrrr@{}}
\toprule
Dataset & Model & Seed & Val student & Test student & Random F & Selective F & $\Delta$ [95\% CI] \\
\midrule
ECF & Qwen & 42 & 66.06 & 65.04 & 65.17 & 66.73 & +1.56 [+0.76, +2.37] \\
ECF & Qwen & 43 & 65.45 & 64.86 & 65.02 & 66.23 & +1.21 [+0.38, +2.03] \\
ECF & Qwen & 44 & 65.91 & 64.95 & 65.14 & 66.49 & +1.35 [+0.55, +2.18] \\
ECF & Gemma & 42 & 64.40 & 63.25 & 63.52 & 64.75 & +1.23 [+0.37, +2.17] \\
ECF & Gemma & 43 & 64.76 & 64.01 & 64.30 & 66.34 & +2.04 [+1.18, +2.93] \\
ECF & Gemma & 44 & 64.25 & 64.12 & 64.60 & 66.97 & +2.37 [+1.55, +3.20] \\
MEC$^4$ & Qwen & 42 & 47.70 & 49.83 & 49.89 & 50.26 & +0.38 [-0.36, +1.14] \\
MEC$^4$ & Qwen & 43 & 47.92 & 49.93 & 49.98 & 50.56 & +0.58 [-0.12, +1.25] \\
MEC$^4$ & Qwen & 44 & 47.38 & 49.87 & 49.89 & 50.26 & +0.36 [-0.41, +1.16] \\
MEC$^4$ & Gemma & 42 & 46.80 & 49.00 & 49.15 & 50.35 & +1.20 [+0.37, +1.98] \\
MEC$^4$ & Gemma & 43 & 46.38 & 48.95 & 49.05 & 50.03 & +0.98 [+0.30, +1.64] \\
MEC$^4$ & Gemma & 44 & 46.77 & 49.34 & 49.49 & 50.65 & +1.17 [+0.52, +1.82] \\
\bottomrule
\end{tabular}
\end{table*}

\section{Expert and Calibration Robustness}
\label{app:merged-K}
\label{sec:s-robust}
\subsection{Cross-family expert responses}
\label{app:merged-K-1}
Table~\ref{tab:s-cross-family} changes the expert while retaining the student's query set. Qwen students receive the Gemma expert, and Gemma students receive Qwen. Keeping the requests fixed isolates the effect of different returned judgments under fusion and replacement.

On ECF, switching Qwen students to Gemma improves fusion by 0.56 F1 at 10\%, while switching Gemma students to Qwen reduces it by 0.61. Both paired intervals exclude zero. MEC$^4$ changes are smaller, with intervals including zero. The precision--recall profiles in Table~\ref{tab:s-experts} help explain why experts with similar overall F1 can differ in their usefulness.

\begin{table*}[!tbp]
\centering
\caption{Experts from another family, mean$_{\scriptscriptstyle\pm\mathrm{SD}}$ test F1. Student identifies the querying model. Cross uses the other family's expert with student queries unchanged. $\Delta$ is Cross F minus Same F.}
\label{tab:s-cross-family}
\begin{tabular}{@{}llrrrrr@{}}
\toprule
Dataset & Student & $b$ (\%) & Same F & Cross F & Cross R & $\Delta$ [95\% CI] \\
\midrule
ECF & Qwen & 5 & $63.69_{\scriptscriptstyle\pm0.63}$ & $64.04_{\scriptscriptstyle\pm0.30}$ & $64.19_{\scriptscriptstyle\pm0.35}$ & +0.35 [+0.02, +0.70] \\
ECF & Qwen & 10 & $64.89_{\scriptscriptstyle\pm0.79}$ & $65.45_{\scriptscriptstyle\pm0.34}$ & $65.54_{\scriptscriptstyle\pm0.53}$ & +0.56 [+0.02, +1.14] \\
ECF & Qwen & 20 & $65.51_{\scriptscriptstyle\pm0.80}$ & $66.16_{\scriptscriptstyle\pm0.35}$ & $66.69_{\scriptscriptstyle\pm0.14}$ & +0.66 [+0.03, +1.32] \\
ECF & Gemma & 5 & $63.95_{\scriptscriptstyle\pm0.20}$ & $63.49_{\scriptscriptstyle\pm0.37}$ & $63.86_{\scriptscriptstyle\pm0.12}$ & -0.46 [-0.86, -0.07] \\
ECF & Gemma & 10 & $65.14_{\scriptscriptstyle\pm0.29}$ & $64.53_{\scriptscriptstyle\pm0.44}$ & $65.01_{\scriptscriptstyle\pm0.21}$ & -0.61 [-1.18, -0.05] \\
ECF & Gemma & 20 & $65.99_{\scriptscriptstyle\pm0.36}$ & $65.17_{\scriptscriptstyle\pm0.45}$ & $66.33_{\scriptscriptstyle\pm0.24}$ & -0.82 [-1.54, -0.07] \\
MEC$^4$ & Qwen & 5 & $49.10_{\scriptscriptstyle\pm0.74}$ & $49.09_{\scriptscriptstyle\pm0.89}$ & $49.30_{\scriptscriptstyle\pm0.63}$ & -0.02 [-0.51, +0.45] \\
MEC$^4$ & Qwen & 10 & $49.74_{\scriptscriptstyle\pm0.60}$ & $49.66_{\scriptscriptstyle\pm0.76}$ & $50.22_{\scriptscriptstyle\pm0.21}$ & -0.08 [-0.69, +0.51] \\
MEC$^4$ & Qwen & 20 & $49.90_{\scriptscriptstyle\pm0.54}$ & $50.02_{\scriptscriptstyle\pm0.63}$ & $50.24_{\scriptscriptstyle\pm0.04}$ & +0.12 [-0.55, +0.74] \\
MEC$^4$ & Gemma & 5 & $48.08_{\scriptscriptstyle\pm0.20}$ & $48.00_{\scriptscriptstyle\pm0.17}$ & $48.15_{\scriptscriptstyle\pm0.20}$ & -0.08 [-0.52, +0.36] \\
MEC$^4$ & Gemma & 10 & $49.02_{\scriptscriptstyle\pm0.86}$ & $49.06_{\scriptscriptstyle\pm0.45}$ & $49.39_{\scriptscriptstyle\pm0.15}$ & +0.04 [-0.55, +0.66] \\
MEC$^4$ & Gemma & 20 & $49.45_{\scriptscriptstyle\pm1.01}$ & $49.37_{\scriptscriptstyle\pm0.61}$ & $49.85_{\scriptscriptstyle\pm0.10}$ & -0.08 [-0.71, +0.59] \\
\bottomrule
\end{tabular}
\end{table*}

\subsection{Resampling the calibration dialogues}
\label{app:merged-K-2}
Each student receives 200 bootstrap samples of validation dialogues. Every draw refits both affine calibrators, decision thresholds, and query bands before evaluation. All draws contribute to the sensitivity summary. Table~\ref{tab:s-cal-sensitivity} reports the gain distribution and actual query range. Its central ranges describe variation in the calibration sample, separately from the paired intervals over evaluation dialogues.

Consultation improves over the corresponding recalibrated student in every resample. The smallest gain is 1.70 F1 for Gemma/ECF seed 44. Both query membership and query count change with recalibration, yet the gain persists.

\begin{table*}[!tbp]
\centering
\caption{Calibration sensitivity at nominal 10\% access. Columns report consultation gain over the recalibrated student, its central range and minimum, the fraction of positive draws, and the actual access range.}
\label{tab:s-cal-sensitivity}
\begin{tabular}{@{}llrrrrrr@{}}
\toprule
Dataset & Model & Seed & Mean $\Delta$ & 2.5--97.5\% & Minimum & Positive (\%) & Query range (\%) \\
\midrule
ECF & Qwen & 42 & +2.62 & [+2.20, +3.27] & +1.95 & 100 & [8.62, 10.72] \\
ECF & Qwen & 43 & +3.01 & [+2.37, +3.58] & +1.86 & 100 & [8.24, 10.21] \\
ECF & Qwen & 44 & +2.82 & [+2.12, +3.74] & +2.07 & 100 & [8.59, 10.48] \\
ECF & Gemma & 42 & +3.87 & [+3.19, +4.38] & +2.86 & 100 & [8.19, 10.38] \\
ECF & Gemma & 43 & +3.67 & [+2.89, +4.48] & +2.59 & 100 & [8.14, 10.22] \\
ECF & Gemma & 44 & +3.64 & [+2.29, +4.53] & +1.70 & 100 & [8.44, 10.73] \\
MEC$^4$ & Qwen & 42 & +2.50 & [+2.16, +2.77] & +1.90 & 100 & [9.42, 10.82] \\
MEC$^4$ & Qwen & 43 & +2.77 & [+2.47, +2.97] & +2.24 & 100 & [9.58, 10.45] \\
MEC$^4$ & Qwen & 44 & +2.64 & [+2.26, +3.43] & +1.92 & 100 & [9.24, 10.62] \\
MEC$^4$ & Gemma & 42 & +3.15 & [+2.51, +3.83] & +2.48 & 100 & [9.18, 10.57] \\
MEC$^4$ & Gemma & 43 & +4.67 & [+4.04, +5.89] & +3.89 & 100 & [9.40, 10.49] \\
MEC$^4$ & Gemma & 44 & +2.83 & [+2.35, +3.37] & +1.75 & 100 & [9.37, 10.66] \\
\bottomrule
\end{tabular}
\end{table*}

\section{How Gains Change with Query Allowance}
\label{app:merged-L}
\label{sec:s-budgets}
The following tables retain every student seed at nominal 5/10/20\% access. $b$ and $q$ denote nominal and actual query percentages. Stud., Rnd., and Sel. denote the student, matched random queries, and selective queries. F and R denote fusion and replacement. All 36 student--budget configurations improve validation F1 with selective fusion.

\subsection{Qwen on ECF}
\label{app:merged-L-1}
For Qwen/ECF, selective fusion F1 increases with access for all three seeds on both splits (Tables~\ref{tab:s-qwen_ecf-val} and~\ref{tab:s-qwen_ecf-test}). Selective fusion also exceeds matched random consultation at every test budget.

Replacement benefits more from the wider 20\% band, reaching test F1 of 66.27--66.36 versus 64.59--66.05 for fusion, and 66.78--67.21 on validation.

\begin{table*}[!tbp]
\centering
\begin{minipage}[t]{0.48\textwidth}
\centering
\caption{Qwen / ECF on validation, pair F1.}
\label{tab:s-qwen_ecf-val}
\begin{tabular}{@{}rrrrrrrr@{}}
\toprule
Seed & $b$ & $q$ & Stud. & Rnd. F & Sel. F & Rnd. R & Sel. R \\
\midrule
42 & 5 & 4.29 & 63.05 & 63.24 & 63.86 & 63.27 & 64.23 \\
42 & 10 & 9.64 & 63.05 & 63.31 & 65.32 & 63.47 & 65.37 \\
42 & 20 & 19.66 & 63.05 & 63.52 & 65.64 & 64.04 & 66.78 \\
43 & 5 & 4.79 & 63.12 & 63.24 & 64.75 & 63.34 & 64.97 \\
43 & 10 & 9.72 & 63.12 & 63.46 & 65.31 & 63.60 & 65.74 \\
43 & 20 & 19.56 & 63.12 & 63.75 & 65.80 & 64.04 & 67.21 \\
44 & 5 & 4.11 & 63.82 & 64.02 & 64.72 & 64.08 & 65.28 \\
44 & 10 & 9.48 & 63.82 & 64.04 & 65.52 & 64.20 & 66.41 \\
44 & 20 & 19.42 & 63.82 & 64.43 & 66.22 & 64.76 & 66.89 \\
\bottomrule
\relax

\end{tabular}
\end{minipage}\hfill
\begin{minipage}[t]{0.48\textwidth}
\centering
\caption{Qwen / ECF on test, pair F1.}
\label{tab:s-qwen_ecf-test}
\begin{tabular}{@{}rrrrrrrr@{}}
\toprule
Seed & $b$ & $q$ & Stud. & Rnd. F & Sel. F & Rnd. R & Sel. R \\
\midrule
42 & 5 & 4.21 & 61.28 & 61.46 & 62.97 & 61.60 & 63.26 \\
42 & 10 & 9.64 & 61.28 & 61.60 & 64.00 & 61.86 & 64.75 \\
42 & 20 & 19.46 & 61.28 & 61.93 & 64.59 & 62.36 & 66.36 \\
43 & 5 & 4.53 & 62.04 & 62.27 & 63.98 & 62.28 & 64.11 \\
43 & 10 & 9.03 & 62.04 & 62.46 & 65.11 & 62.52 & 65.41 \\
43 & 20 & 19.62 & 62.04 & 62.93 & 65.87 & 63.02 & 66.35 \\
44 & 5 & 3.64 & 62.82 & 62.95 & 64.13 & 62.98 & 64.38 \\
44 & 10 & 9.24 & 62.82 & 63.18 & 65.55 & 63.18 & 66.03 \\
44 & 20 & 19.14 & 62.82 & 63.58 & 66.05 & 63.69 & 66.27 \\
\bottomrule
\relax

\end{tabular}
\end{minipage}
\end{table*}

\subsection{Gemma on ECF}
\label{app:merged-L-2}
Gemma/ECF also benefits beyond the nearest 5\% band (Tables~\ref{tab:s-gemma_ecf-val} and~\ref{tab:s-gemma_ecf-test}). Both response rules improve every student at every allowance. At 20\%, replacement reaches 66.40--67.01 on validation and exceeds fusion for each seed.

The test results retain the advantage over the student and random access. Fusion and replacement are close at 10\%, while replacement is stronger at 20\%.

\begin{table*}[!tbp]
\centering
\begin{minipage}[t]{0.48\textwidth}
\centering
\caption{Gemma / ECF on validation, pair F1.}
\label{tab:s-gemma_ecf-val}
\begin{tabular}{@{}rrrrrrrr@{}}
\toprule
Seed & $b$ & $q$ & Stud. & Rnd. F & Sel. F & Rnd. R & Sel. R \\
\midrule
42 & 5 & 4.97 & 61.78 & 61.94 & 63.79 & 62.16 & 63.98 \\
42 & 10 & 9.30 & 61.78 & 62.07 & 64.52 & 62.31 & 65.20 \\
42 & 20 & 19.84 & 61.78 & 62.46 & 64.96 & 62.86 & 66.40 \\
43 & 5 & 4.47 & 62.25 & 62.41 & 64.16 & 62.49 & 64.36 \\
43 & 10 & 9.23 & 62.25 & 62.76 & 66.08 & 62.80 & 65.77 \\
43 & 20 & 19.26 & 62.25 & 63.18 & 66.43 & 63.39 & 67.01 \\
44 & 5 & 4.53 & 61.63 & 61.96 & 63.71 & 62.06 & 63.77 \\
44 & 10 & 9.75 & 61.63 & 62.27 & 65.76 & 62.51 & 65.88 \\
44 & 20 & 19.88 & 61.63 & 62.63 & 66.15 & 62.89 & 66.82 \\
\bottomrule
\relax

\end{tabular}
\end{minipage}\hfill
\begin{minipage}[t]{0.48\textwidth}
\centering
\caption{Gemma / ECF on test, pair F1.}
\label{tab:s-gemma_ecf-test}
\begin{tabular}{@{}rrrrrrrr@{}}
\toprule
Seed & $b$ & $q$ & Stud. & Rnd. F & Sel. F & Rnd. R & Sel. R \\
\midrule
42 & 5 & 4.75 & 61.35 & 61.60 & 64.00 & 61.60 & 64.01 \\
42 & 10 & 8.92 & 61.35 & 61.84 & 64.88 & 61.95 & 65.18 \\
42 & 20 & 19.66 & 61.35 & 62.26 & 65.69 & 62.28 & 66.28 \\
43 & 5 & 4.18 & 61.87 & 62.05 & 64.13 & 62.07 & 64.04 \\
43 & 10 & 8.71 & 61.87 & 62.35 & 65.46 & 62.37 & 65.27 \\
43 & 20 & 18.81 & 61.87 & 62.79 & 66.38 & 62.81 & 66.65 \\
44 & 5 & 4.45 & 60.56 & 60.83 & 63.73 & 60.90 & 63.68 \\
44 & 10 & 9.61 & 60.56 & 61.16 & 65.09 & 61.25 & 65.16 \\
44 & 20 & 19.49 & 60.56 & 61.68 & 65.89 & 61.98 & 66.37 \\
\bottomrule
\relax

\end{tabular}
\end{minipage}
\end{table*}

\subsection{Qwen on \texorpdfstring{MEC$^4$}{MEC4}}
\label{app:merged-L-3}
Qwen has a flatter fusion curve, with most of the test improvement already present at 10\% (Tables~\ref{tab:s-qwen_mec4-val} and~\ref{tab:s-qwen_mec4-test}). For seed 42, raising access to 20\% changes fusion from 50.27 to 50.32 but lowers replacement from 50.15 to 49.75. The extra answers mostly leave fusion decisions unchanged, while replacement can overturn correct student judgments.

The 20\% test bands yield 20.19--20.80\% actual access. A band transferred from validation admits a fraction determined by the new score distribution. The global comparisons in Section~\ref{sec:s-crossed} use a hard answer count instead.

\begin{table*}[!tbp]
\centering
\begin{minipage}[t]{0.48\textwidth}
\centering
\caption{Qwen / MEC$^4$ on validation, pair F1.}
\label{tab:s-qwen_mec4-val}
\begin{tabular}{@{}rrrrrrrr@{}}
\toprule
Seed & $b$ & $q$ & Stud. & Rnd. F & Sel. F & Rnd. R & Sel. R \\
\midrule
42 & 5 & 4.92 & 45.18 & 45.43 & 48.01 & 45.36 & 47.63 \\
42 & 10 & 9.76 & 45.18 & 45.63 & 49.17 & 45.61 & 49.14 \\
42 & 20 & 19.08 & 45.18 & 46.00 & 49.25 & 45.83 & 48.74 \\
43 & 5 & 4.89 & 45.30 & 45.44 & 46.72 & 45.48 & 47.21 \\
43 & 10 & 9.34 & 45.30 & 45.64 & 47.77 & 45.73 & 48.45 \\
43 & 20 & 19.74 & 45.30 & 45.87 & 48.02 & 46.02 & 48.59 \\
44 & 5 & 4.55 & 43.59 & 43.83 & 45.77 & 43.83 & 45.73 \\
44 & 10 & 8.82 & 43.59 & 44.05 & 47.08 & 44.13 & 47.53 \\
44 & 20 & 19.94 & 43.59 & 44.57 & 47.62 & 44.63 & 48.66 \\
\bottomrule
\relax

\end{tabular}
\end{minipage}\hfill
\begin{minipage}[t]{0.48\textwidth}
\centering
\caption{Qwen / MEC$^4$ on test, pair F1.}
\label{tab:s-qwen_mec4-test}
\begin{tabular}{@{}rrrrrrrr@{}}
\toprule
Seed & $b$ & $q$ & Stud. & Rnd. F & Sel. F & Rnd. R & Sel. R \\
\midrule
42 & 5 & 5.14 & 47.74 & 47.91 & 49.82 & 47.89 & 49.68 \\
42 & 10 & 10.14 & 47.74 & 48.07 & 50.27 & 48.06 & 50.15 \\
42 & 20 & 20.19 & 47.74 & 48.28 & 50.32 & 48.24 & 49.75 \\
43 & 5 & 4.92 & 47.10 & 47.27 & 49.14 & 47.32 & 49.00 \\
43 & 10 & 9.58 & 47.10 & 47.45 & 49.87 & 47.49 & 49.98 \\
43 & 20 & 20.65 & 47.10 & 47.78 & 50.10 & 47.75 & 49.92 \\
44 & 5 & 4.81 & 46.46 & 46.69 & 48.35 & 46.67 & 48.75 \\
44 & 10 & 9.24 & 46.46 & 46.81 & 49.09 & 46.83 & 49.56 \\
44 & 20 & 20.80 & 46.46 & 47.17 & 49.29 & 47.34 & 49.90 \\
\bottomrule
\relax

\end{tabular}
\end{minipage}
\end{table*}

\subsection{Gemma on \texorpdfstring{MEC$^4$}{MEC4}}
\label{app:merged-L-4}
At wider access, replacement varies less across Gemma students (Tables~\ref{tab:s-gemma_mec4-val} and~\ref{tab:s-gemma_mec4-test}). At 20\%, it ranges from 48.44 to 48.49 on validation and 50.19--50.27 on test. Test fusion ranges from 48.42 to 50.44. Replacing selected decisions with a shared expert reduces the influence of differences among student scores.

With each student's query set held fixed, integration can affect variation across students. Lower variation need not imply a larger gain, as the repair and damage counts in Section~\ref{sec:s-revisions} illustrate.

\begin{table*}[!tbp]
\centering
\begin{minipage}[t]{0.48\textwidth}
\centering
\caption{Gemma / MEC$^4$ on validation, pair F1.}
\label{tab:s-gemma_mec4-val}
\begin{tabular}{@{}rrrrrrrr@{}}
\toprule
Seed & $b$ & $q$ & Stud. & Rnd. F & Sel. F & Rnd. R & Sel. R \\
\midrule
42 & 5 & 4.51 & 43.44 & 43.57 & 45.51 & 43.64 & 45.68 \\
42 & 10 & 9.85 & 43.44 & 43.82 & 46.65 & 43.97 & 47.09 \\
42 & 20 & 19.78 & 43.44 & 44.11 & 46.96 & 44.44 & 48.49 \\
43 & 5 & 4.44 & 43.24 & 43.43 & 45.84 & 43.43 & 45.95 \\
43 & 10 & 9.07 & 43.24 & 43.63 & 47.21 & 43.69 & 47.23 \\
43 & 20 & 19.07 & 43.24 & 44.14 & 47.83 & 44.26 & 48.44 \\
44 & 5 & 4.59 & 43.26 & 43.45 & 45.86 & 43.46 & 45.17 \\
44 & 10 & 9.20 & 43.26 & 43.75 & 46.97 & 43.75 & 47.03 \\
44 & 20 & 19.63 & 43.26 & 44.11 & 47.28 & 44.23 & 48.44 \\
\bottomrule
\relax

\end{tabular}
\end{minipage}\hfill
\begin{minipage}[t]{0.48\textwidth}
\centering
\caption{Gemma / MEC$^4$ on test, pair F1.}
\label{tab:s-gemma_mec4-test}
\begin{tabular}{@{}rrrrrrrr@{}}
\toprule
Seed & $b$ & $q$ & Stud. & Rnd. F & Sel. F & Rnd. R & Sel. R \\
\midrule
42 & 5 & 4.62 & 45.65 & 45.80 & 48.12 & 45.78 & 48.62 \\
42 & 10 & 10.10 & 45.65 & 46.02 & 49.18 & 46.05 & 49.86 \\
42 & 20 & 20.34 & 45.65 & 46.44 & 49.48 & 46.47 & 50.27 \\
43 & 5 & 4.55 & 45.14 & 45.37 & 48.27 & 45.38 & 48.32 \\
43 & 10 & 9.40 & 45.14 & 45.74 & 49.80 & 45.68 & 49.76 \\
43 & 20 & 19.70 & 45.14 & 46.02 & 50.44 & 45.85 & 50.23 \\
44 & 5 & 4.59 & 45.71 & 45.86 & 47.87 & 45.92 & 48.01 \\
44 & 10 & 9.46 & 45.71 & 46.01 & 48.10 & 46.14 & 49.20 \\
44 & 20 & 20.32 & 45.71 & 46.26 & 48.42 & 46.50 & 50.19 \\
\bottomrule
\relax

\end{tabular}
\end{minipage}
\end{table*}

\section{Reconstructing the Reported Results}
\label{app:merged-M}
The supplementary material provides selected consultation and evaluation code with frozen relation scores, candidate indices, and complete gold pairs. The primary CSV records 72 results by setting, seed, split, and budget. A separate CSV records candidate coverage, decision probabilities, and query composition for Table~\ref{tab:s-query-composition}. Calibration files retain affine coefficients, thresholds, and bands. Other records include competing policies, learned priorities, integration parameters, paired intervals, and the four continued students. Scores and gold pairs also support reconstruction of emotion and cause endpoint metrics.

The supplied script reconstructs query masks, predictions, F1, and paired intervals, checking them against the stored results. Queries reveal expert responses only at selected positions, and unqueried decisions remain unchanged. Zero queries or using the student as expert reproduce its decisions. An optional routine refits the learned routers from validation features. Supporting records document KD, calibration resampling, cross-family consultation, and GPU timing. The complete training and native inference code will be released upon acceptance.

\makeatletter
\if@firstcolumn\newpage\fi
\makeatother
\subsection{Use of AI Assistants}\label{app:ai-use}
\label{app:merged-M-1}
The artificial intelligence (AI) assistant GPT Codex 5.6 sol was used to implement and debug experimental code and produce plots from experimental results. Writing assistance primarily involved language polishing and condensation of the authors' text, with limited rewriting of selected passages based on material they supplied. The authors reviewed the text and code produced with AI assistance and verified reported results against experimental outputs. The authors made the final methodological decisions and take full responsibility for the scientific claims and the accuracy and integrity of the manuscript and accompanying artifacts.

\end{document}